\pdfoutput=1
\documentclass[11pt]{article}

\usepackage[final]{acl}

\usepackage{times}
\usepackage{latexsym}

\usepackage[T1]{fontenc}
\usepackage[utf8]{inputenc}

\usepackage{microtype}

\usepackage{inconsolata}

\usepackage{graphicx}

\usepackage{amsmath}
\usepackage{amssymb}
\usepackage{mathtools}
\usepackage{amsthm}
\usepackage[capitalize,noabbrev]{cleveref}
\usepackage[textsize=tiny]{todonotes}
\usepackage{xspace}
\newcommand{\name}{ProBench\xspace}
\newcommand{\eg}{\emph{e.g.}}
\newcommand{\ie}{i.\,e.}
\usepackage{url}
\usepackage{hyperref}
\usepackage[capitalize]{cleveref}
\crefname{figure}{Fig.}{Figs.}
\Crefname{figure}{Figure}{Figures}
\crefname{section}{Sec.}{Secs.}
\Crefname{section}{Section}{Sections}
\Crefname{table}{Table}{Tables}
\crefname{table}{Tab.}{Tabs.}
\Crefname{equation}{Equation}{Equations}
\crefname{equation}{Eq.}{Eqs.}
\usepackage{pgfplots}
\usepackage{booktabs}
\usepackage{tabularx}
\newcolumntype{Y}{>{\centering\arraybackslash}X}
\usepackage{multicol}
\usepackage{multirow}
\usepackage{fontawesome}
\usepackage{colortbl}
\definecolor{deemph}{gray}{0.6}
\newcommand{\gc}[1]{\textcolor{deemph}{#1}} 
\usepackage{makecell}
\usepgfplotslibrary{statistics}

\newcommand{\ariaemoji}{\includegraphics[height=1.2\fontcharht\font`\B, width=1.2\fontcharht\font`\B, width=1.2\fontcharht\font`\B]{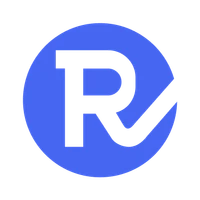}\xspace}
\newcommand{\googleemoji}{\includegraphics[height=1.2\fontcharht\font`\B, width=1.2\fontcharht\font`\B, width=1.2\fontcharht\font`\B]{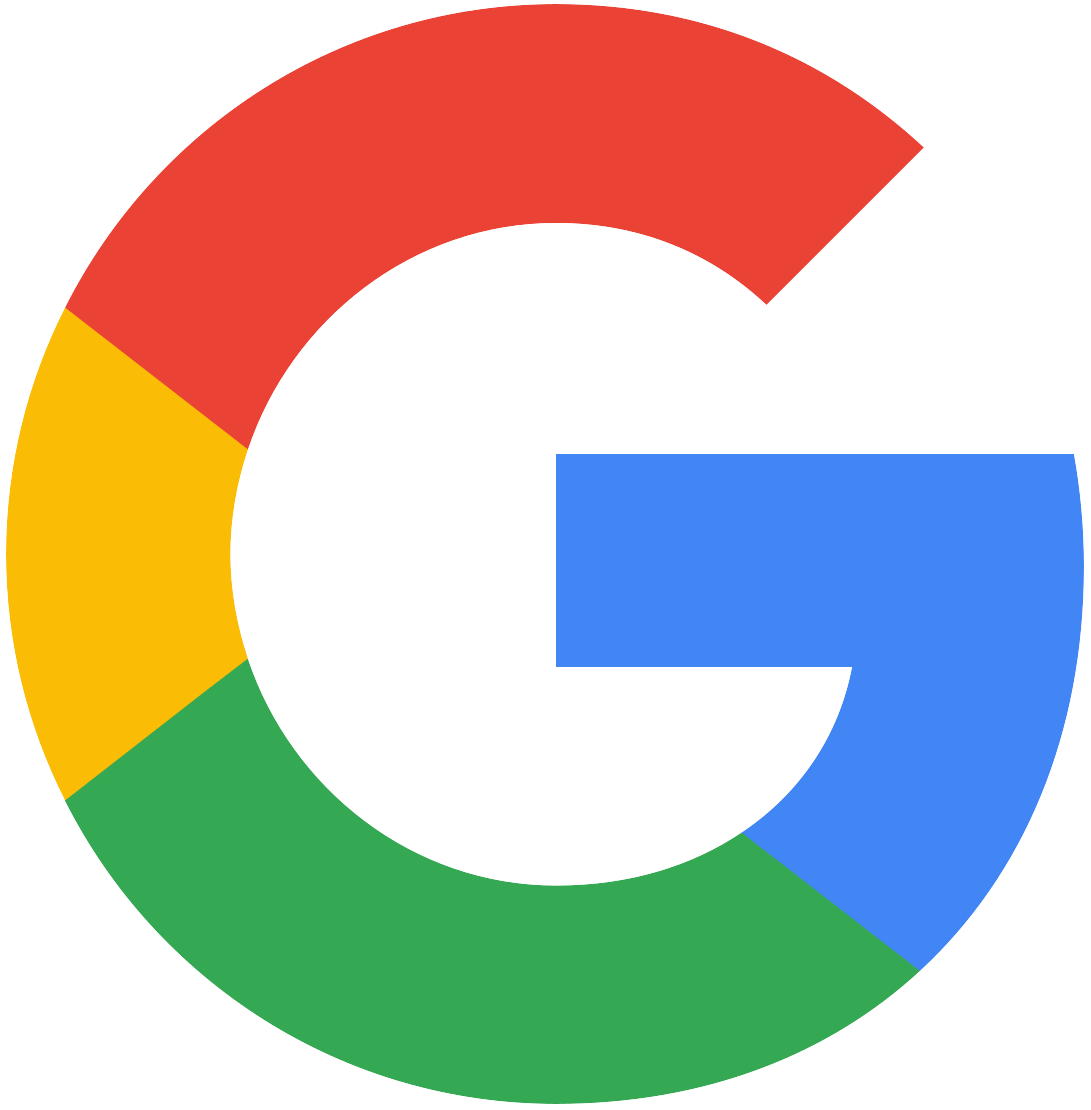}\xspace}
\newcommand{\openaiemoji}{\includegraphics[height=1.2\fontcharht\font`\B, width=1.2\fontcharht\font`\B]{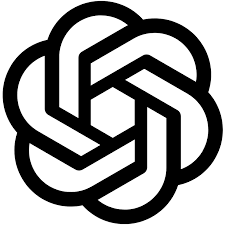}\xspace}
\newcommand{\claudeemoji}{\includegraphics[height=1.2\fontcharht\font`\B, width=1.2\fontcharht\font`\B]{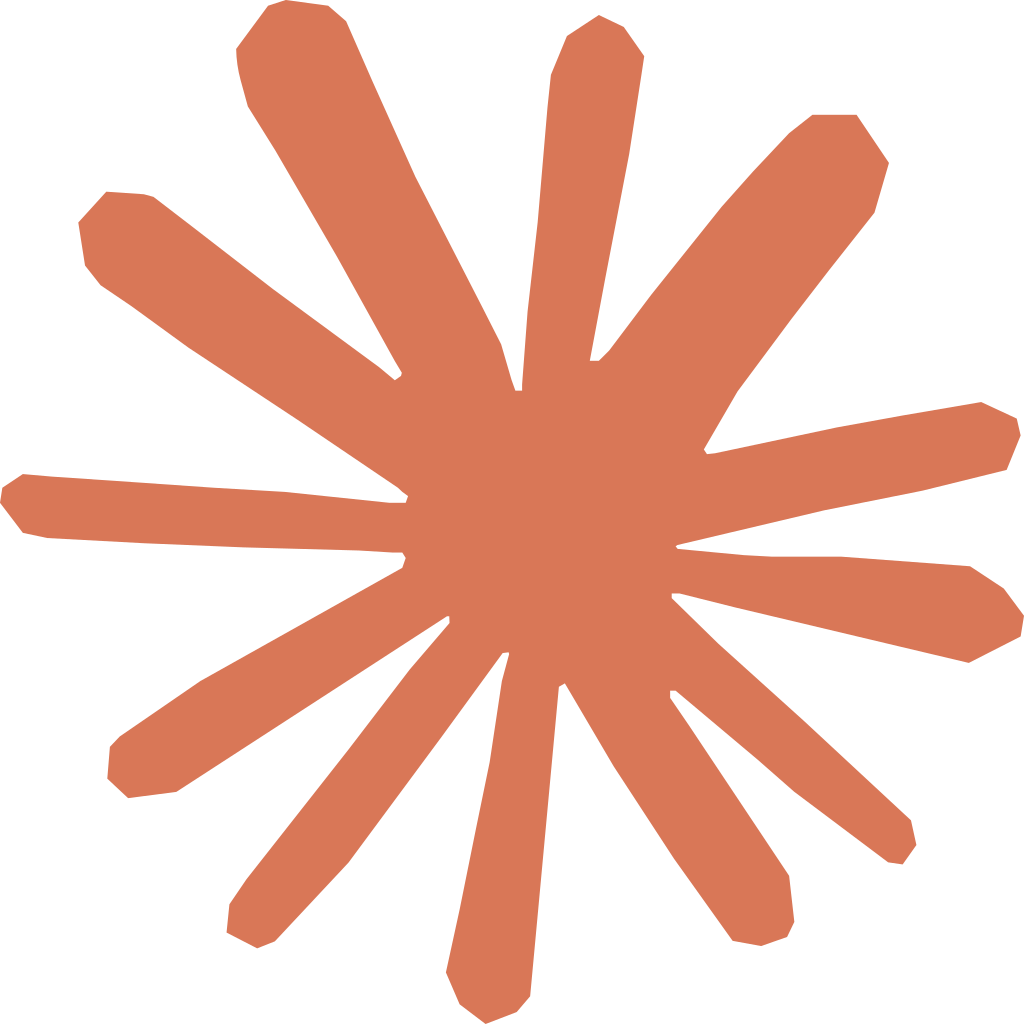}\xspace}
\newcommand{\qwenemoji}{\includegraphics[height=1.2\fontcharht\font`\B, width=1.2\fontcharht\font`\B]{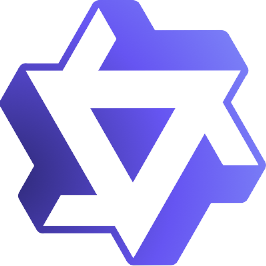}\xspace}
\newcommand{\molmoemoji}{\includegraphics[height=1.2\fontcharht\font`\B, width=1.2\fontcharht\font`\B]{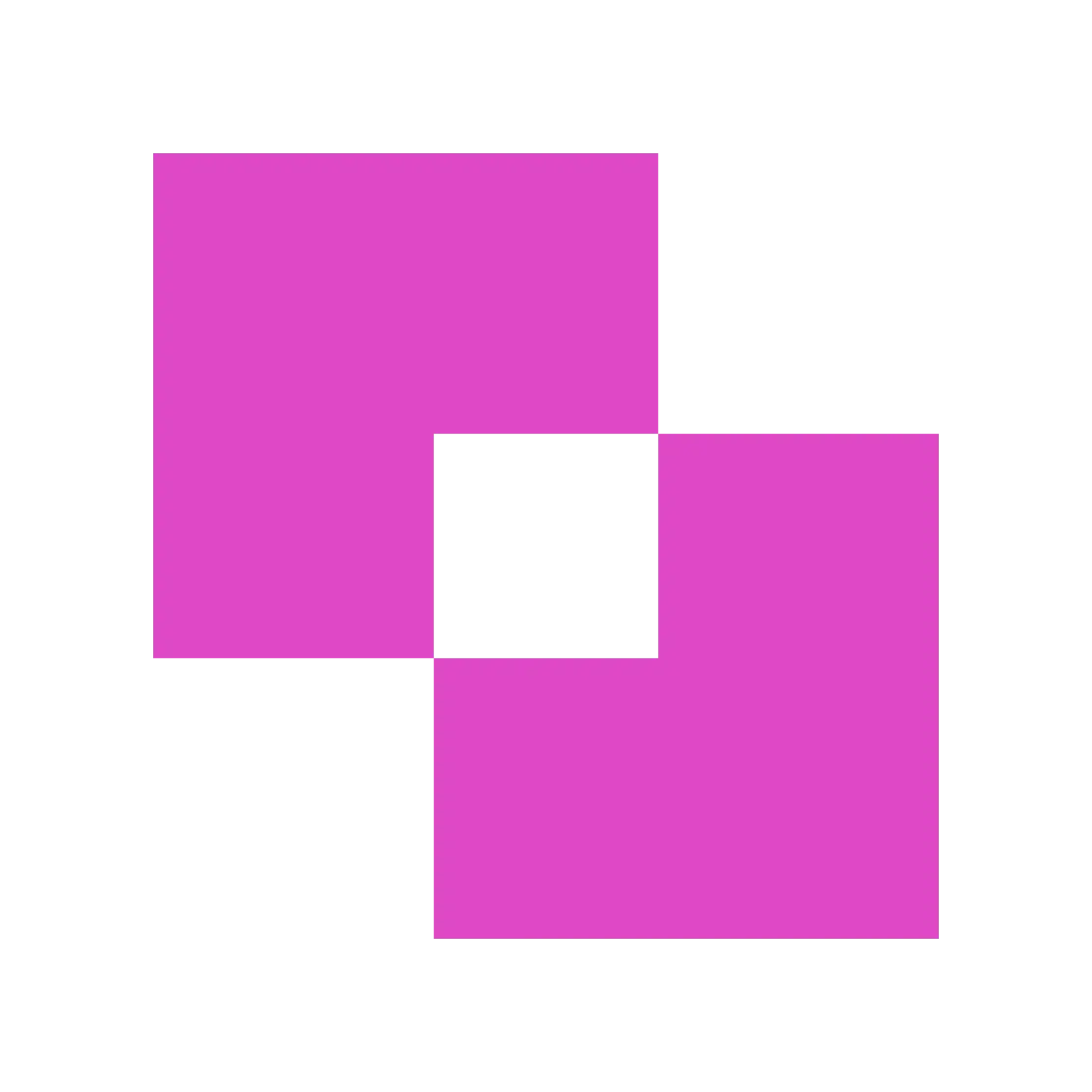}\xspace}
\newcommand{\internvlemoji}{\includegraphics[height=1.2\fontcharht\font`\B, width=1.2\fontcharht\font`\B]{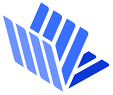}\xspace}
\newcommand{\minicpmemoji}{\includegraphics[height=1.2\fontcharht\font`\B, width=1.2\fontcharht\font`\B]{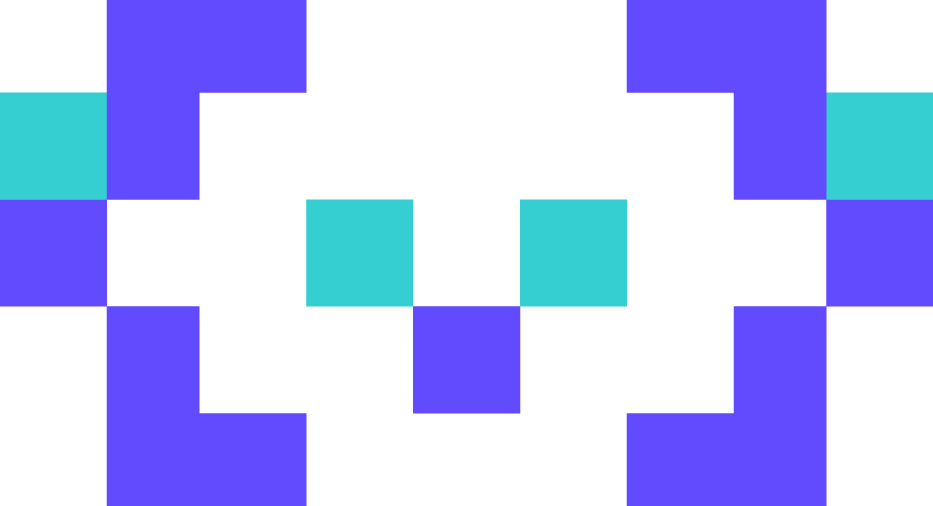}\xspace}
\newcommand{\llavaemoji}{\includegraphics[height=1.2\fontcharht\font`\B, width=1.2\fontcharht\font`\B]{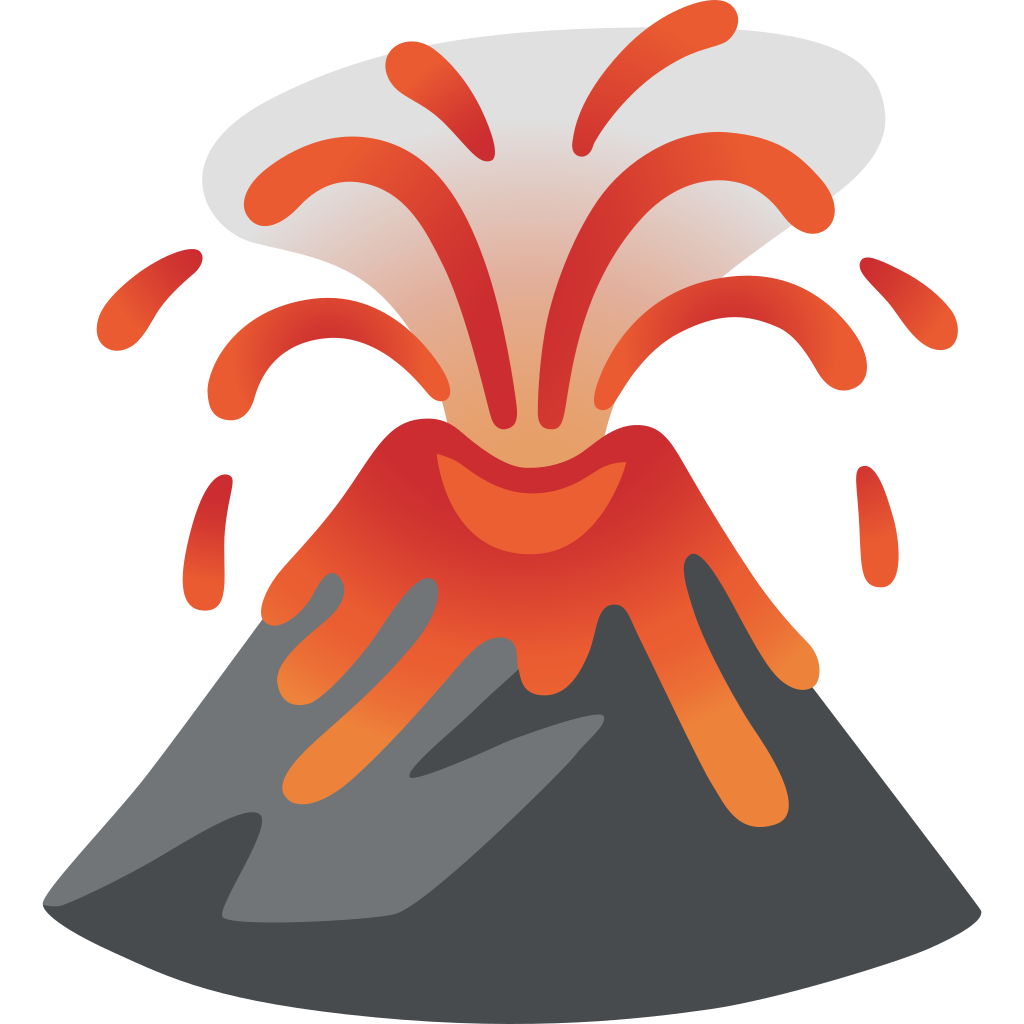}\xspace}
\newcommand{\llamaemoji}{\includegraphics[height=1.2\fontcharht\font`\B, width=1.2\fontcharht\font`\B]{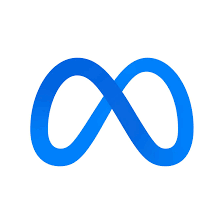}\xspace}
\newcommand{\pixtralemoji}{\includegraphics[height=1.2\fontcharht\font`\B, width=1.2\fontcharht\font`\B]{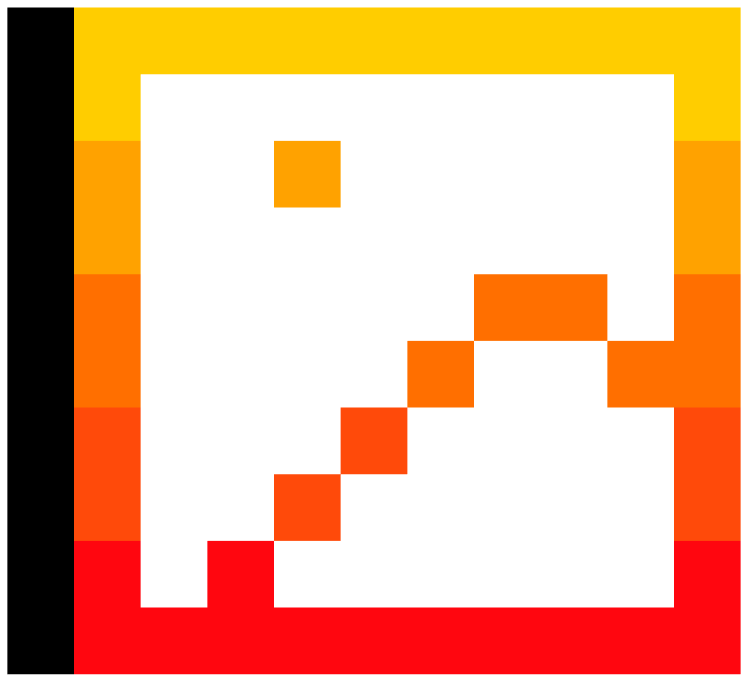}\xspace}
\newcommand{\nvlmemoji}{\includegraphics[height=1.2\fontcharht\font`\B, width=1.2\fontcharht\font`\B]{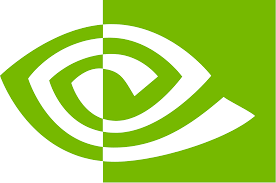}\xspace}

\pgfplotsset{compat=1.18}
\usepackage{subfig}
\usepackage{caption}
\usepackage{pgf-pie}
\usepackage{pgfplotstable}
\definecolor{forestgreen}{RGB}{69,137,51}
\usepackage{subcaption}

\usepackage[most]{tcolorbox}
\tcbuselibrary{skins}
\tcbuselibrary{breakable}
\usepackage[smartEllipses]{markdown}
\usepackage{listings}
\usepackage{listingsutf8}
\usepackage{longtable}
\UseRawInputEncoding
\usepackage[T1]{fontenc}
\usepackage[utf8]{inputenc} 
\usepackage[export]{adjustbox}
\usepackage{fancyvrb} 
\usepackage{cuted} 
\usepackage{bbding}

\title{ProBench: Judging Multimodal Foundation Models on \\ Open-ended Multi-domain Expert Tasks}

\author{
    Yan Yang$^1$ \quad Dongxu Li$^1$ \Envelope \quad Haoning Wu$^2$ \quad Bei Chen \quad Liu Liu$^3$ \quad Liyuan Pan$^4$ \Envelope \quad Junnan Li$^5$ \\
    $^1$ANU \quad $^2$NTU \quad $^3$KooMap, Huawei \quad $^4$BITSZ \& School of CSAT, BIT \quad $^5$Salesforce AI Research \\
  \texttt{dongxuli1005@gmail.com} \quad  \texttt{liyuan.pan@bit.edu.cn} \\
  Project Page: \url{https://yan98.github.io/ProBench/}}

\begin{document}
    \maketitle

\begin{strip}
  \centering
  \includegraphics[width=\textwidth]{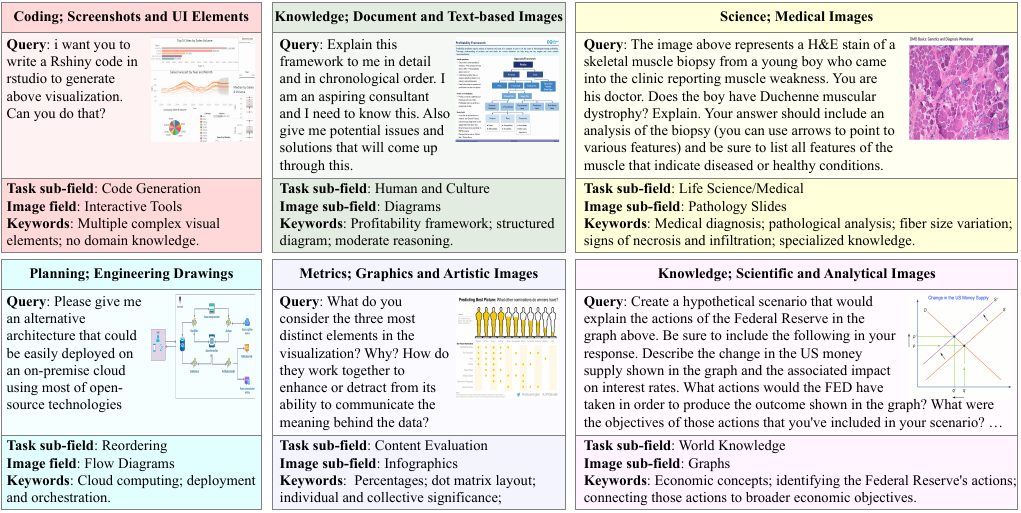}
  \vspace{-2em}
  \captionof{figure}{\small Examples of \name. 
  Our \name spans 10 task fields and 56 sub-fields, supports 17 languages, and supports conversations with up to 13 conversation turns. We show the task and image fields in the header of each sample. We use `Engineering Drawings' for `Engineering and Technical Drawings' in the first plot of the second row due to space constraints. More diverse and longer samples are provided in the supplementary material.}
  \label{fig:example}
\end{strip}

\begin{abstract}

Solving expert-level multimodal tasks is a key milestone towards general intelligence. As the capabilities of multimodal large language models (MLLMs) continue to improve, evaluation of such advanced multimodal intelligence becomes necessary yet challenging. In this work, we introduce~\name, a benchmark of open-ended user queries that require professional expertise and advanced reasoning. \name consists of 4,000 high-quality samples independently submitted by professionals based on their daily productivity demands. It spans across 10 fields and 56 sub-fields, including science, arts, humanities, coding, mathematics, and creative writing. Experimentally, we evaluate and compare 24 latest models using MLLM-as-a-Judge. Our results reveal that although the best open-source models rival the proprietary ones, \name presents significant challenges in visual perception, textual understanding, domain knowledge and advanced reasoning, thus providing valuable directions for future multimodal AI research efforts.
\end{abstract}

\section{Introduction}

Solving expert-level multimodal tasks with multimodal  large language models (MLLMs) represents an important milestone toward achieving human-level general intelligence. However, these tasks require accurate user query understanding, in-depth domain-specific knowledge, and advanced reasoning abilities, which present significant challenges for frontier models as of today. Measuring such progress requires rigorous evaluations. To this end, we introduce \name, a challenging and automatic evaluation benchmark leveraging MLLM-as-a-Judge. \name consists of 4,000 queries submitted independently by professional users, covering diverse productivity demands and expert knowledge to assess MLLM capabilities in open-ended scenarios (\cref{fig:example}).

One common benchmark to evaluate MLLM performance with expert knowledge is MMMU \cite{yue2024mmmu}. While effective for automatic evaluation using predefined answer choices, such benchmarks fail to capture MLLM capabilities in open-ended user interactions. Specifically, they do not adequately assess MLLM ability to follow user instructions or align with human preferences. Both are fundamental aspects for real-world applications \cite{lu2024wildvision, luo2024videoautoarena, chen2024mega}. Similar limitations apply to other benchmarks, such as MMMU-pro \cite{yue2024mmmupro}, MMBench \cite{liu2025mmbench}, among others \cite{lu2023mathvista, masry2022chartqa, singh2019towards, wu2024longvideobench}.

\begin{figure}[!t]
    \centering
    \includegraphics{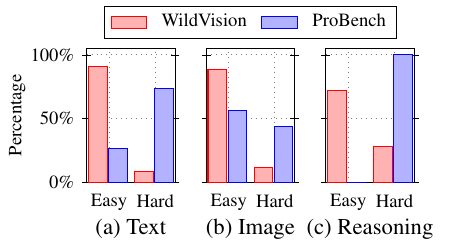}  
    \vspace{-2.51em}
    \caption{\small Comparison with WildVision \cite{lu2024wildvision} on challenge levels of (a) text, (b) image, and (c) reasoning for user instruction queries. To ensure a fair comparison, we follow WildVision by selecting the top 500 highest-quality queries from the single-round conversations. It can be seen that \name contains significantly more hard samples than WildVision.
    }
    \label{fig:compare}
\end{figure}

Alternatively, MLLM-as-a-Judge is usually employed to automatically evaluate model performance in open-ended scenarios. However, existing open-ended multimodal benchmarks require limited expert-level or professional knowledge. Among them, some \cite{chen2024mega} are constructed by few experts, limiting their domain coverage, while remaining ones \cite{luo2024videoautoarena, lu2024wildvision}, such as WildVision, are mostly set in general chat environments and require much less domain knowledge to solve.

\begin{figure}[!t]
    \centering
    \includegraphics{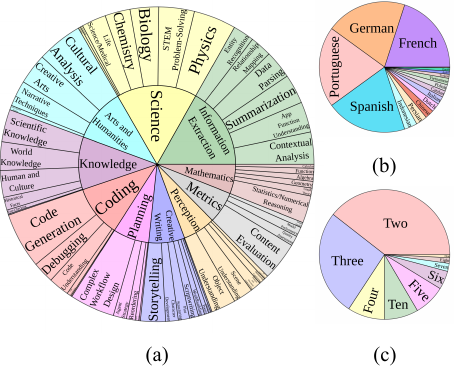}
    \vspace{-2em}
    \caption{\small \name overview. Distributions of (a) task fields on the single-round track, (b) languages  on the multi-linguistic track, and (c) conversation rounds on the multi-round tracks. 
    }
    \label{fig:overview}
\end{figure}

To fill this gap, in this paper, we aim to design an \textit{open-ended benchmark that requires expert-level knowledge} for multimodal tasks. Our \name is created from high-quality interactions within 100K real-world, professionally crowdsourced multimodal conversations for productivity scenarios. Specifically, samples are collected by encouraging professionals to ask questions related to their daily professional work, which usually require significant expert-level knowledge. This distinction sets our benchmark apart from prior works like WildVision \cite{lu2024wildvision} (\cref{fig:compare}). For a comprehensive evaluation, \name includes three tracks: single-round, multi-round, and multi-linguistic conversations. They respectively span 10 task fields and 56 sub-fields, support 17 languages, and support conversations with up to 13 conversation turns. An overview of \name is presented in \cref{fig:overview}.

\begin{figure*}[!t]
    \centering
    \includegraphics{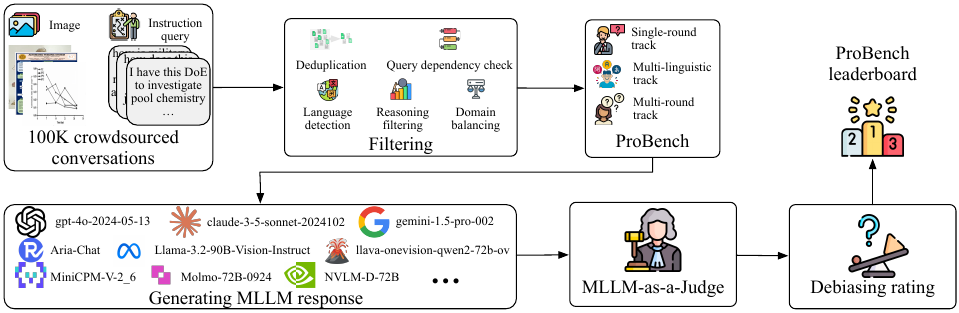}
    \vspace{-1.5em}
    \caption{\small Framework of \name. Starting with 100K crowdsourced conversations, we identify high-quality user queries to curate single-round, multi-linguistic, and multi-round tracks. Using MLLM-as-a-Judge, we benchmark and rank 24 state-of-the-art MLLMs with ELO ratings. To ensure fairness, the ELO ratings are de-biased to remove confounder effects (\eg, MLLM response formats), resulting in the final \name leaderboard. Icons in the figure are sourced from \cite{flaticon2025}. }
    \label{fig:method}
\end{figure*}

Leveraging MLLM-as-a-Judge (\eg, gpt-4o), we assess 24 leading MLLMs on \name.  Our evaluation reveals several key limitations in state-of-the-art MLLMs:
i) current MLLMs struggle in visual perception, textual understanding, domain knowledge, and advanced reasoning, suffering from tasks like mathematics and planning;
ii) multi-linguistic understanding and long-context reasoning during multi-round interaction remain challenging for most existing MLLMs. Our main contributions are summarized as follows:
\begin{itemize}
    \vspace{-.5em}
    \setlength\itemsep{-.5em}
    \item we introduce~\name, an open-ended multimodal benchmark tailored for professional work scenarios requiring expert-level knowledge, featuring 4,000 samples across 10 task fields over 56 sub-fields. The benchmark also features multi-round conversations up to 13 turns and multi-linguistic tracks in 17 languages;
    \item we design an automatic pairwise evaluation pipeline using MLLM-as-a-Judge, achieving 79.9\% agreement with human experts. The evaluation is robust to different comparison baseline and judge model choices. We also provide a distilled version of Llama-vision to support cost-effective local evaluations;
    \item we conduct comprehensive evaluations on 24 leading MLLMs, showing that~\name presents significant challenges for existing MLLMs, in visual perception, advanced reasoning, and domain knowledge. This signifies the need for more advanced multimodal models for high-value practical scenarios.
\end{itemize}

\section{\name}
\paragraph{Preliminary.} The \name dynamically ranks MLLMs by employing the ELO rating system, implemented through statistical modeling based on direct pairwise model comparisons. In the following, we provide an overview. For further details, please refer to \cite{elo1966uscf, hunter2004mm}. Given $N$ MLLMs, an online ELO rating system compares model $i$ with rating $\boldsymbol{r}_{i}$ and model $j$ with rating $\boldsymbol{r}_{j}$ using the probability $P(\boldsymbol{y}_{i,j} = 1)$. Here, $\boldsymbol{y}_{i,j}$ denotes the binary outcome, where $\boldsymbol{y}_{i,j} = 1$ indicates that model $i$ wins, and $\boldsymbol{y}_{i,j} = 0$ indicates that model $j$ wins. The probability is calculated by 
\begin{equation*}
    P(\boldsymbol{y}_{i,j} = 1) = \frac{1}{1 + 10^{(\boldsymbol{r}_{i} - \boldsymbol{r}_{j}) / \alpha}} \ ,
\end{equation*}
where $\alpha$ is a hyperparameter that serves as a scaling factor, typically set to $\alpha = 400$. The ELO rating is dynamically updated after each model comparison. Taking model $i$ as an example, the rating is updated according to the following rule:
\begin{equation*}
    \boldsymbol{r}^{\mathsf{upt}}_{i} = \boldsymbol{r}_{i}  + K \times (\boldsymbol{s}_{i,j} - P(\boldsymbol{y}_{i,j} = 1) ) \ .
\end{equation*}
Similarly, $K$ is a constant determining the magnitude of rating adjustments, commonly set to $K = 32$. The term $\boldsymbol{s}_{i,j}$ is a scalar representing the actual outcome: $0$ for a loss, $0.5$ for a tie, and $1$ for a win. This updating rule encourages that a higher-rated model gains fewer points for a win, and loses more points for a defeat, while a lower-rated model experiences the opposite effect. 

However, when using MLLM-as-a-Judge, the comparison results can be sensitive to model presentation order and confounded by response style variations \cite{li2024crowdsourced}. To address these challenges, the \name incorporates the Bradley-Terry model \cite{hunter2004mm} as an additional layer atop the ELO system. For $N$ MLLMs and $M$ pairwise comparisons, each round $1 \leq m \leq M$ compares model $i$ and model $j$. We have $\mathbf{X}^{\mathsf{win}}_m \in \mathbb{R}^{N}$ to indicate which model is presented first\footnote{This bias can be easily mitigated by evaluating twice while swapping the comparison order.},
while $\mathbf{X}^{\mathsf{sty}}_m \in \mathbb{R}^{S}$ captures $S$ stylistic differences between the outputs of models $i$ and $j$ (\eg, word counts, and use of markdown). The Bradley-Terry model then refines the rating of model $i$ as
\begin{align*}
    \boldsymbol{r}^{\mathsf{ref}}_{i} &= C + K \times \hat{\beta}_{i} \ , \\
    \hat{\beta}, \hat{\gamma} &= \arg \min_{\beta, \gamma} \sum_{m,i,j} \ell_{\mathsf{bce}}(\beta^\top \mathbf{X}^{\mathsf{win}}_{m} + \gamma^\top \mathbf{X}^{\mathsf{sty}}_{m}, \boldsymbol{s}_{i,j}) \ ,
\end{align*}
where $\ell_{\mathsf{bce}}(\cdot, \cdot)$ is the binary cross-entropy loss,  $C$ is a baseline rating constant, $\beta \in \mathbb{R}^{N}$ and $\gamma \in \mathbb{R}^{S}$ are respectively known as the model strength and style coefficients, and $\hat{\beta}_{i}$ is a scaler indicating strength of model $i$. This refinement known as style control in the literature \cite{li2024does} compensates for stylistic biases, ensuring a fair model performance evaluation.

\paragraph{Overview.} We aim to establish a comprehensive and challenging benchmark for evaluating MLLMs. The resulting \name is built on two primary components:
i) curating high-quality conversations from a crowdsourced data, categorized into single-round,  multi-linguistic, and multi-round tracks;
ii) employing MLLM-as-a-Judge to compare and rank MLLMs.
In total, 3000, 500, and 500 conversations are selected for the single-round,  multi-linguistic, and multi-round tracks, respectively, from an initial pool of 100K crowdsourced user-MLLM conversations. An overview is presented in \cref{fig:method}.

\subsection{Benchmark establishment}
The benchmark is curated based on three guiding principles:  
i) diversity, selected user instruction queries target to avoid redundancies while extensively covering MLLM-based tasks;
ii) MLLM-driven, the chosen queries of conversations are tailored to evaluate the unique capabilities of MLLMs in the multimodal domain;
iii) coherence, the benchmark enables targeted evaluations for specific MLLM tasks, rather than providing undifferentiated evaluations. We first describe the common steps involved in curating the three tracks, followed by a discussion of the track-specific methodologies.

\paragraph{Common step.} We filter out short user instruction queries that contain excessive stop words, and apply MinHash-based text deduplication \cite{lee2021deduplicating} to retain a pool of non-redundant queries. To address potential redundancy or irrelevance between the instructions and images within a user query, we perform image-instruction deduplication. This step removes queries that can be sufficiently answered using only the textual instructions, leveraging an MLLM-based filter.

\paragraph{Single-round track.} A language detector is employed to filter out non-English user instruction queries. Starting with a pool of MLLM task and sub-task fields derived from \cite{chen2024mega}, we use an MLLM-based annotator to assign user instruction queries to existing fields or propose new ones where necessary. Additionally, the annotator assesses the challenge level of each query. To ensure diversity, domain balancing is performed, and overrepresented task fields are downsampled, resulting in 3000 user instruction queries.

\paragraph{Multi-linguistic track.} User instruction queries are categorized by their languages, excluding all English-based conversations. Based on frequency, the queries are grouped into Portuguese (PT), French (FR), Spanish (ES), German (DE), and an ``Other'' category (\eg, Chinese, Vietnamese, and more). An MLLM-based annotator is then used to assess the challenges of the queries, with the 100 most difficult queries retained for each group.

\paragraph{Multi-round track.} Similar to the single-round track, we focus on user instruction queries in English for this track. Multi-round conversations are required to feature interconnected queries across rounds, demonstrating a progressive nature. To achieve this, we identify the reasoning challenges and interdependencies between queries within the conversations, applying an MLLM annotator. Ultimately, the 100 most challenging independent queries and 400 interconnected multi-round user instruction queries are preserved.

Detailed prompts used for the above steps are provided in the supplementary material. With the \name, we are readily to assess and rank the MLLMs. 

\begin{figure*}[p]
    \centering
    \makebox[\textwidth][c]{\includegraphics{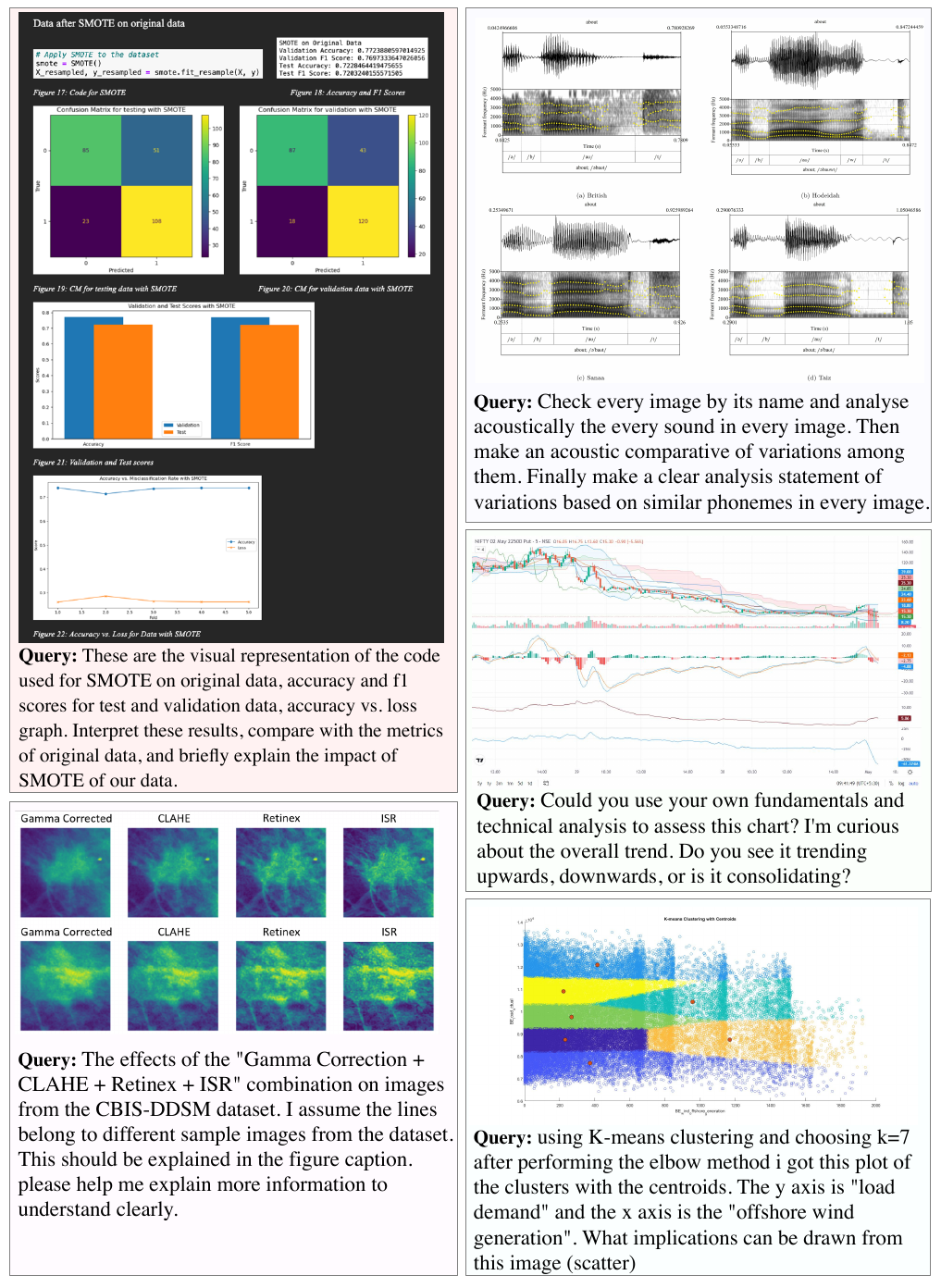}}
    \caption{Example queries from \name. As shown, significant domain knowledge and reasoning capabilities are needed to solve \name queries. For brevity, we only show examples with relatively shorter text queries, with the remark that longer queries are common in \name. More examples can be found in the appendix.}
    \label{fig:example_question}
\end{figure*}

\begin{table*}[!t]
    \centering
    \scriptsize
    \caption{\small Comparisons of state-of-the-art MLLMs on the single-round track are presented using the following abbreviations: Sci. (Science), Cd. (Coding), CW. (Creative Writing), IE. (Information Extraction), Perc. (Perception), Knowl. (Knowledge), Arts (Arts), Plan. (Planning), Math (Mathematics), and Mt. (Metrics). We provide ELO ratings for each task, followed by an overview that includes the average number of output tokens (\#Token), 95\% confidence interval (95\% CI), win rate (WR), and overall ELO rating. The MLLMs are sorted by the overal ELo rating in each group of model size. }
    \vspace{-1em}
    \setlength{\tabcolsep}{3.2pt}
    \begin{tabularx}{\linewidth}{lr|cccccccccc|cccc}
    \toprule
    \multirow{2}{*}{\bf Model} & & \multicolumn{10}{c|}{\bf Task-Specific ELO Ratings} & \multicolumn{4}{c}{\bf Overview} \\
    \cmidrule{3-16}
    & & Sci. & Cd. & CW. & IE. & Perc. & Knowl. & Arts & Plan. & Math. & Mt. & \#Token & 95\% CI & WR & Elo \\
    \midrule
    \multicolumn{2}{l|}{\textit{\textbf{\gc{Proprietary MLLMs}}}} & \multicolumn{10}{l|}{} \\ 
    \claudeemoji claude-3-5-sonnet-20241022    & \faLock{} & 1228  & 1252  & 1259  & 1211  & 1213  & 1272  & 1236  & 1192  & 1197  & 1251  & 405 &   (-7, 8)    & 65.84 & 1228\\
    \googleemoji gemini-1.5-pro-002 & \faLock{} & 1151  & 1145  & 1105  & 1100  & 1110  & 1067  & 1107  & 1095  & 1134  & 1147  & 500 &   (-8, 10)   & 50.58 & 1118\\ 
    \openaiemoji gpt-4o-2024-05-13 & \faLock{} & 1114  & 1114  & 1114  & 1114  & 1114  & 1114  & 1114  & 1114  & 1114  & 1114  & 491 &    (0, 0)    & 50.00 & 1114 \\
    \openaiemoji gpt-4o-mini-2024-07-18 & \faLock{} & 1049  & 1074  & 1165  & 1094  & 1096  & 1101  & 1130  & 1102  & 1037  & 1159  & 526 &   (-8, 10)   & 47.12 & 1094 \\
    \openaiemoji gpt-4o-2024-08-06 & \faLock{} & 1096  & 1112  & 1050  & 1097  &  995  & 1080  & 1032  & 1058  & 1175  & 1015  & 374 &   (-7, 7)    & 44.98 & 1079\\
    \googleemoji gemini-1.5-flash-002  & \faLock{} & 1025  &  877  & 1092  & 1007  & 1022  & 1011  &  993  &  946  & 1035  & 1087  & 493 &   (-8, 9)    & 35.33 & 1009 \\
    \midrule
    \multicolumn{2}{l|}{\textit{\textbf{\gc{70B+ Open-source MLLMs}}}} & \multicolumn{10}{l|}{} \\
    \pixtralemoji Pixtral-Large-Instruct-2411 & 124B & 1230  & 1194  & 1280  & 1242  & 1224  & 1250  & 1245  & 1221  & 1175  & 1266  & 715 &   (-8, 8)    & 65.97 & 1229 \\
    \internvlemoji InternVL2\_5-78B & 78B & 1083  & 1018  & 1051  & 1091  & 1031  & 1084  & 1042  & 1073  & 1065  & 1023  & 558 &   (-7, 10)   & 42.85 & 1064 \\
    \qwenemoji Qwen2-VL-72B-Instruct & 72B & 1009  &  914  &  965  &  991  &  986  &  960  &  962  &  921  &  998  &  970  & 557 &   (-9, 9)    & 31.37 &  978 \\
    \molmoemoji Molmo-72B-0924 & 72B & 828  &  733  &  953  &  859  &  903  &  881  &  862  &  817  &  871  &  852  & 301 &   (-12, 8)   & 18.46 &  856 \\
    \nvlmemoji NVLM-D-72B & 72B & 780  &  877  &  991  &  810  &  849  &  835  &  767  &  881  &  838  &  725  & 561 &  (-10, 10)   & 16.63 &  834 \\
    \llamaemoji Llama-3.2-90B-Vision-Instruct & 90B &  830  &  751  &  624  &  754  &  806  &  842  &  626  &  769  &  940  &  662  & 448 &  (-11, 10)   & 12.89 &  782 \\
    \llavaemoji llava-onevision-qwen2-72b-ov & 72B & 696  &  735  &  762  &  726  &  767  &  689  &  663  &  679  &  853  &  620  & 360 &  (-11, 12)   & 10.09 &  734 \\
    \midrule
    \multicolumn{2}{l|}{\textit{\textbf{\gc{10B+ Open-source MLLMs}}}} & \multicolumn{10}{l|}{} \\
    \pixtralemoji Pixtral-12B-2409 & 12B & 1028  &  965  & 1099  & 1031  & 1024  & 1057  & 1047  & 1083  &  996  & 1063  & 659 &   (-5, 8)    & 39.1 & 1037 \\
    \ariaemoji Aria-Chat & 3.9/25.3B& 990  &  982  &  985  &  937  &  998  & 1034  & 1019  &  974  &  973  & 1016  & 675 &   (-7, 8)    & 32.88 &  990 \\
    \internvlemoji InternVL2\_5-38B & 38B & 1000  &  979  & 1028  &  987  & 1021  &  904  &  932  & 1041  & 1026  &  933  & 521 &   (-9, 9)    & 32.5 &  987 \\
    \internvlemoji InternVL2\_5-26B & 26B &  890  &  816  & 1008  &  894  &  944  &  876  &  864  &  964  &  880  &  896  & 490 &   (-10, 8)   & 22.59 &  900 \\
    \llamaemoji Llama-3.2-11B-Vision-Instruct & 11B & 671  &  541  &  681  &  702  &  766  &  761  &  624  &  524  &  744  &  614  & 531 &  (-13, 16)   & 7.93 &  688 \\
    \midrule
    \multicolumn{2}{l|}{\textit{\textbf{\gc{7B+ Open-source MLLMs}}}} & \multicolumn{10}{l|}{} \\
    \internvlemoji InternVL2\_5-8B & 8B &  824  &  806  &  983  &  880  &  914  &  840  &  915  &  895  &  835  &  868  & 644 &   (-11, 8)   & 20.45 &  878 \\
    \qwenemoji Qwen2-VL-7B-Instruct & 7B &  803  &  689  &  827  &  877  &  861  &  816  &  736  &  680  &  858  &  833  & 787 &   (-9, 10)   & 15.40 &  818 \\
    \minicpmemoji MiniCPM-V-2\_6 & 8B  &  644  &  599  &  767  &  659  &  812  &  676  &  673  &  667  &  656  &  681  & 646 &  (-12, 10)   & 7.97 &  689 \\
    \llavaemoji llava-onevision-qwen2-7b-ov & 7B &  605  &  570  &  807  &  683  &  809  &  681  &  715  &  608  &  573  &  724  & 575 &  (-13, 10)   & 7.93 &  688 \\
    \molmoemoji Molmo-7B-D-0924 & 7B &  536  &  304  &  720  &  631  &  638  &  655  &  681  &  531  &  613  &  603  & 310 &  (-14, 12)   & 5.41 &  617 \\
    \molmoemoji Molmo-7B-O-0924 & 7B &  457  &  134  &  623  &  483  &  681  &  599  &  606  &  380  &  428  &  528  & 296 &  (-18, 19)   & 3.54 &  540\\
    \bottomrule
    \end{tabularx}
    \label{tab:singleround}
\end{table*}

\subsection{MLLM-as-a-Judge and ranking}  
We evaluate MLLM performance in addressing user instruction queries using a 5-point Likert scale \cite{likert1932technique}, by conducting pairwise comparisons against a baseline model (\eg, GPT-4o). While evaluations by domain-specific human experts are considered as the gold standard, they are resource-intensive, time-consuming, and challenging to scale for large-scale benchmarks. As an alternative, we employ MLLM-as-a-Judge as an approximation of human expertise \cite{li2024crowdsourced,zheng2023judging,chen2024mllm}. The MLLM-as-a-Judge is guided by the following principles.
\begin{itemize}
    \vspace{-0.5em}
    \setlength\itemsep{-0.3em}
    \item \textbf{Correctness:} ensures the accuracy of information, absence of factual errors, and alignments with known and visual knowledge. (For the multi-linguistic track, response language consistency is emphasized).
    \item \textbf{Helpfulness:} provides clear, practical, and actionable guidance to address the user instruction query.
    \item \textbf{Relevance:} focuses on the prompt requirements, avoiding extraneous or tangential information.
    \item \textbf{Conciseness:} avoids unnecessary verbosity while maintaining clarity and direct language.
    \item \textbf{Completeness:} covers all essential aspects of the user instruction query, providing sufficient information to address it.
    \vspace{-0.5em}
\end{itemize}
Details of the prompts used to guide MLLM-as-a-Judge are provided in the supplementary material. Subsequently, we apply the ELO rating system, as described in the preliminary section, to compute the de-biased ratings of each MLLM. These ratings are used for leaderboard comparisons, ensuring a fair and consistent evaluation across models.

\section{Experiment}

\subsection{Experimental setup}
\paragraph{Implementation detail.} All MLLMs are benchmarked using the \texttt{vllm} \cite{kwon2023efficient} and \texttt{HuggingFace} \cite{wolf2019huggingface} codebases, with greedy sampling employed for response generation. For MLLMs with limited context lengths (\eg, a 4096 token context in Molmo-7B-D-0924), sliding window generation is applied to handle longer inputs.  Our MLLM judge utilizes gpt-4o-2024-08-06 with greedy sampling for consistent and reproducible evaluation. For pairwise comparisons in Elo rating calculations,  we set gpt-4o-2024-05-13 as the baseline, evaluate each model twice by swapping the presentation order for each user query, and de-bias the ELO ratings by following the methodology of \cite{li2024crowdsourced}.

\vspace{-3mm}
\paragraph{MLLM.} We evaluate 24 leading MLLMs: gpt-4o-mini-2024-07-18 \cite{hurst2024gpt}, gpt-4o-2024-08-06 \cite{hurst2024gpt}, gpt-4o-2024-05-13 \cite{hurst2024gpt}, claude-3-5-sonnet-20241022   \cite{anthropic2024claude}, gemini-1.5-pro-002 \cite{team2023gemini}, gemini-1.5-flash-002 \cite{team2023gemini}, Aria-Chat \cite{li2024aria}, InternVL2\_5-8B \cite{wang2024enhancing}, InternVL2\_5-26B \cite{wang2024enhancing}, InternVL2\_5-38B \cite{wang2024enhancing}, InternVL2\_5-78B \cite{wang2024enhancing}, Pixtral-12B-2409 \cite{agrawal2024pixtral}, Pixtral-Large-Instruct-2411 \cite{agrawal2024pixtral}, Qwen2-VL-7B-Instruct \cite{wang2024qwen2}, Qwen2-VL-72B-Instruct \cite{wang2024qwen2}, MiniCPM-V-2\_6 \cite{yao2024minicpm},  Llama-3.2-11B-Vision-Instruct \cite{dubey2024llama}, Llama-3.2-90B-Vision-Instruct \cite{dubey2024llama}, Molmo-7B-O-0924 \cite{deitke2024molmo}, Molmo-7B-D-0924 \cite{deitke2024molmo}, Molmo-72B-0924 \cite{deitke2024molmo}, NVLM-D-72B \cite{dai2024nvlm}, llava-onevision-qwen2-7b-ov \cite{li2024llava}, and llava-onevision-qwen2-72b-ov \cite{li2024llava}.

\begin{table*}[!t]
    \centering
    \scriptsize
    \caption{\small Comparisons of state-of-the-art MLLMs on the multi-linguistic and multi-round tracks. We provide an overview that shows the average number of output tokens (\#Token), 95\% confidence interval (95\% CI), win rate (WR), and overall ELO rating for each of the track. Refer to our supplementary material for comparison details on different languages and rounds. The MLLMs are sorted by the overall ELo rating on the multi-linguistic track in each group of model size.
    }
    \vspace{-1em}
    \setlength{\tabcolsep}{3pt}
    \begin{tabularx}{\linewidth}{lr|YYYY|YYYY}
    \toprule
    \multirow{2}{*}{\bf Model} & & \multicolumn{4}{c|}{\bf Overview on multi-linguistic track} & \multicolumn{4}{c}{\bf Overview on multi-round track} \\
    \cmidrule{3-10}
    & &\#Token & 95\% CI & WR & Elo &\#Token & 95\% CI & WR & Elo \\
    \multicolumn{2}{l|}{\textit{\textbf{\gc{Proprietary MLLMs}}}} & \multicolumn{4}{l|}{}\\ 
    \claudeemoji claude-3-5-sonnet-20241022    & \faLock{} & 485 &  (-21, 29)   & 74.58 & 1301 & 1477 &  (-20, 18)   & 70.82 & 1268  \\
    \openaiemoji gpt-4o-2024-05-13 & \faLock{} & 585 &    (0, 0)    & 50.00 & 1114 & 1563 &    (0, 0)    & 50.00 & 1114 \\
    \googleemoji gemini-1.5-pro-002 & \faLock{} & 629 &  (-20, 20)   & 59.11 & 1178 & 1425 &  (-26, 19)   & 53.88 & 1141  \\ 
    \openaiemoji gpt-4o-2024-08-06 & \faLock{}  & 480 &  (-17, 26)   & 60.35 & 1187 & 1052 &  (-22, 18)   & 45.41 & 1082 \\
    \openaiemoji gpt-4o-mini-2024-07-18 & \faLock{} & 657 &  (-21, 16)   & 45.84 & 1085 & 1749 &  (-17, 24)   & 55.16 & 1150  \\
    \googleemoji gemini-1.5-flash-002  & \faLock{} & 567 &  (-25, 19)   & 28.47 &  954 & 1388 &  (-16, 19)   & 38.14 & 1030 \\
    \midrule
    \multicolumn{2}{l|}{\textit{\textbf{\gc{70B+ Open-source MLLMs}}}} & \multicolumn{4}{l|}{} \\
    \pixtralemoji Pixtral-Large-Instruct-2411 & 124B & 966 &  (-23, 22)   & 73.81 & 1294 & 2593 &  (-23, 19)   & 69.73 & 1259  \\
    \qwenemoji Qwen2-VL-72B-Instruct & 72B & 834 &  (-18, 21)   & 47.56 & 1097 & 1608 &  (-21, 19)   & 32.24 &  985 \\
    \internvlemoji InternVL2\_5-78B & 78B & 841 &  (-14, 20)   & 42.71 & 1063 & 2015 &  (-21, 20)   & 44.84 & 1078 \\
    \nvlmemoji NVLM-D-72B & 72B & 907 &  (-17, 25)   & 21.99 &  894 & 1371 &  (-35, 33)   & 8.49 &  701 \\
    \llamaemoji Llama-3.2-90B-Vision-Instruct & 90B & 968 &  (-29, 21)   & 20.92 &  883 & 1350 &  (-36, 24)   & 9.88 &  730 \\
    \molmoemoji Molmo-72B-0924 & 72B & 426 &  (-27, 19)   & 18.90 &  861 & 967 &  (-28, 25)   & 18.64 &  858 \\
    \llavaemoji llava-onevision-qwen2-72b-ov & 72B & 534 &  (-27, 24)   & 11.95 &  767 & 1176 &  (-31, 26)   & 10.30 &  738  \\
    \midrule
    \multicolumn{2}{l|}{\textit{\textbf{\gc{10B+ Open-source MLLMs}}}} & \multicolumn{4}{l|}{} \\
    \internvlemoji InternVL2\_5-38B & 38B  & 868 &  (-20, 18)   & 43.98 & 1072 & 1734 &  (-18, 21)   & 34.68 & 1004 \\
    \pixtralemoji Pixtral-12B-2409 & 12B & 1199 &  (-14, 22)   & 35.73 & 1012 & 2264 &  (-19, 20)   & 40.48 & 1047 \\
    \ariaemoji Aria-Chat & 3.9/25.3B& 1014 &  (-23, 17)   & 35.33 & 1009 & 2321 &  (-27, 12)   & 23.92 &  913 \\
    \internvlemoji InternVL2\_5-26B & 26B & 814 &  (-28, 19)   & 17.70 &  847 & 554 &  (-27, 28)   & 15.77 &  823  \\
    \llamaemoji Llama-3.2-11B-Vision-Instruct & 11B & 2027 &  (-29, 21)   & 8.40 &  699 & 2094 &  (-38, 32)   & 6.03 &  637 \\
    \midrule
    \multicolumn{2}{l|}{\textit{\textbf{\gc{7B+ Open-source MLLMs}}}} & \multicolumn{4}{l|}{} \\
    \qwenemoji Qwen2-VL-7B-Instruct & 7B & 1216 &  (-24, 22)   & 12.25 &  772 & 2004 &  (-34, 25)   & 9.48 &  722 \\
    \internvlemoji InternVL2\_5-8B & 8B & 1021 &  (-22, 20)   & 11.95 &  767 & 1835 &  (-25, 22)   & 11.77 &  764  \\
    \minicpmemoji MiniCPM-V-2\_6 & 8B &  890 &  (-36, 35)   & 4.44 &  581 & 1861 &  (-33, 37)   & 5.35 &  615  \\
    \molmoemoji Molmo-7B-D-0924 & 7B & 406 &  (-52, 33)   & 4.32 &  576 & 923 &  (-34, 26)   & 5.04 &  604 \\
    \llavaemoji llava-onevision-qwen2-7b-ov & 7B & 686 &  (-68, 37)   & 3.07 &  514 & 1743 &  (-30, 30)   & 6.58 &  653 \\
    \molmoemoji Molmo-7B-O-0924 & 7B & 512 &  (-73, 51)   & 1.95 &  433 & 925 &  (-49, 37)   & 3.43 &  534 \\
    \bottomrule
    \end{tabularx}
    \label{tab:multilinguisticmultiround}
\end{table*}

\subsection{Experimental result}
\cref{tab:singleround} and \cref{tab:multilinguisticmultiround} present the evaluation results. Our key observations are summarized into the following five folds:
i) \textit{\textbf{best open-source models rival the best proprietary MLLMs.}} claude-3-5-sonnet-20241022  and Pixtral-Large-Instruct-2411 respectively belonging to proprietary and open-source MLLMs consistently achieve leading ELO scores across all three tracks. Both models significantly outperform the baseline gpt-4o-2024-05-13; 
ii) \textit{\textbf{training recipes make a difference.}} Though scaling parameters can generally improve performance, it is not the sole determining factor. By comparing different models, it shows that training recipes and data quality are also important.
For example, Pixtral with 12B parameters and Aria-Chat with 3.9B activated parameters consistently demonstrate top-tier performance;
iii) \textit{\textbf{reasoning tasks remain the hardest.}}
On the single-round track, most MLLMs generally perform well on writing-based tasks (\eg, creative writing).  
However, their performance on logic-intensive tasks is notably poor, similar to findings in prior LLM studies \cite{ahn2024large, quan2025codeelo}. 
The two tasks separately exhibit the lowest Spearman correlation with overall ELO ratings and receive the lowest scores among task fields. Similarly, among all open-source models, performance also suffers significantly in planning tasks, which have the lowest average score (excluding coding);
iv) \textit{\textbf{multi-linguistic tasks present challenges.}}
MLLMs face significant challenges in multi-linguistic tasks, with 11 out of 24 MLLMs showing an overall ELO decrease compared to their performance on the single-round track. Notably, llava-onevision-qwen2-7b-ov experienced the most substantial decline; v) \textit{\textbf{multi-round evaluations show larger gaps.}}
Multi-round tasks usually demand long-context reasoning across turns, amplifying performance gaps among MLLMs. MLLMs that underperform in single-round tasks exhibit significantly lower ELO scores. This trend is particularly evident in open-source MLLMs with 7B+ and 10B+ parameters (excluding Pixtral-12B-2409).

\begin{figure*}[!t]
    \centering
    \includegraphics{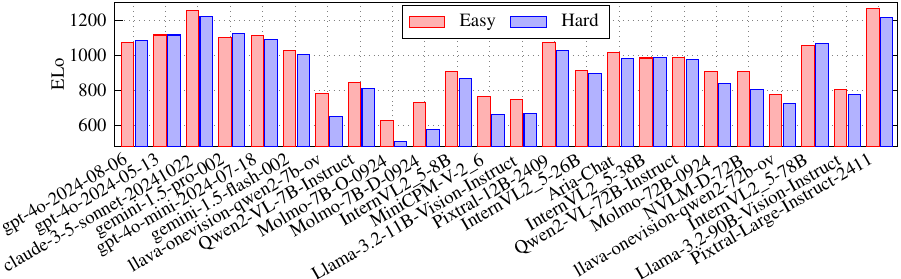}
    \caption{\small Ablation study of reasoning challenge. We show the ELO ratings of MLLMs on  two levels: easy and hard.}
    \label{fig:diffult}
\end{figure*}

\begin{figure*}[!t]
\vspace{-.5em}
\centering
\scriptsize
\includegraphics{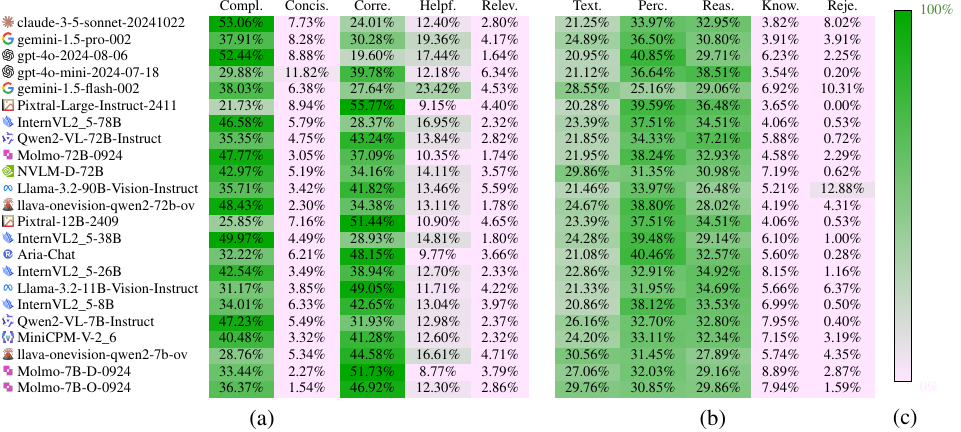}
\vspace{-2em}
\caption{\small Error analysis. We study cases where MLLM underperforms compared to the baseline. (a) The distribution of losing cases of the MLLM across five evaluation aspects: completeness (Compl.), conciseness (Concis.), correctness (Corre.), helpfulness (Helpf.), and relevance (Relve.). (b) The distribution of error types in losses of the MLLM, categorized into five types: textual understanding error (Text.), visual perceptual error (Perc.), reasoning error (Reas.), lack of domain knowledge error (Know.), and refusal to answer (Reje.). (c) Color bar of the heatmap.}
\label{fig:erroranalysis}
\end{figure*}

\subsection{Ablation and discussion}

\paragraph{Performance declining with difficulty.} We evaluate the ELO rating variances of MLLMs by categorizing user queries into easy and hard groups. 
The results are presented in \cref{fig:diffult}.
Existing MLLMs tend to exhibit a noticeable performance decline compared to the baseline gpt-4o-2024-05-13 as the reasoning challenge level increased from easy to hard, while MLLM with poor performance typically deteriorates further on the harder queries. 
This observation aligns with human intuition that more challenging tasks inherently provide better separability when evaluating the MLLM performance, highlighting the limitations of most MLLMs in effectively handling complex user queries.

\vspace{-3mm}
\paragraph{Error analysis.} We analyze scenarios in which the state-of-the-art MLLM underperforms relative to the baseline. \cref{fig:erroranalysis} (a) illustrates the shortcomings of the MLLM compared to the baseline across five evaluation aspects, highlighting completeness and correctness as the primary issues. \cref{fig:erroranalysis} (b) categorizes the error types in the MLLM losses relative to the baseline. Overall, the analysis underscores the need of state-of-the-art MLLM to improve their visual perception, textual understanding, domain knowledge, and reasoning capability.

\begin{figure*}[!t]
    \centering
    \footnotesize
  \includegraphics{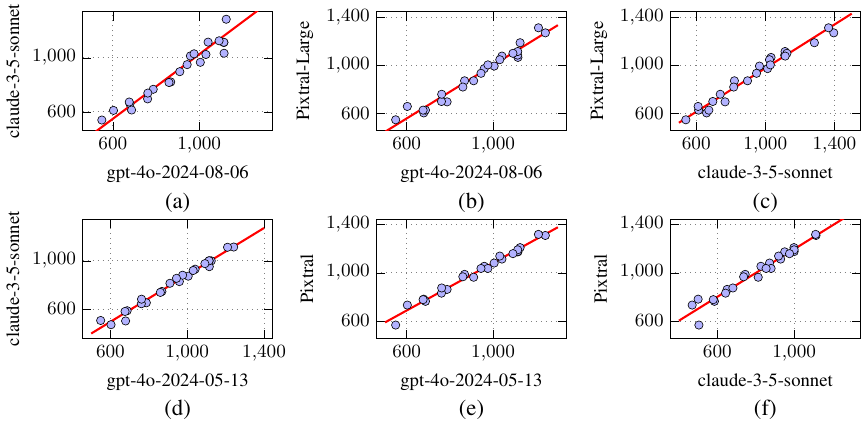}
  \caption{\small Ablation study of MLLM-as-the-Judge. (a-c) Pairwise comparisons of Elo scores for MLLMs evaluated using different MLLM judges. They are gpt-4o-2024-08-06, claude-3-5-sonnet-20241022 (claude-3-5-sonnet), and Pixtral-Large-Instruct-2411 (Pixtral-Large), respectively. (d-f) Comparison of using gpt-4o-2024-05-13, claude-3-5-sonnet-20241022 (claude-3-5-sonnet), and Pixtral-12B-2409 (Pixtral) as baselines. The red line in each plot indicates the best-fit curve for visualization. 
  }
  \label{fig:judge}
\end{figure*}

\vspace{-3mm}
\paragraph{Robustness of \name.} 
We study the setting of our evaluation protocol on the 500 most challenging queries from the single-round track. Specifically, \cref{fig:judge} considers two set of experiments:
i) comparisons of using three top-performing MLLM as the judge (\ie, gpt-4o-2024-08-06, claude-3-5-sonnet-20241022, and Pixtral-Large-Instruct-2411);
ii) explorations of three baseline models (\ie, gpt-4o-2024-05-13, claude-3-5-sonnet-20241022, and Pixtral-12B-2409) in comparisons, representing different model scales. The results reveal a high degree of agreement within our evaluation process, with an average Spearman correlation coefficient of 0.979 among the different MLLM judges and 0.983 among the baseline models, highlighting our robustness and consistency.

\vspace{-3mm}
\paragraph{Judge alignment with human expert.} 
To validate the effectiveness of MLLM-as-a-Judge, human annotators are tasked with rating the comparisons using a 5-point Likert scale. Our evaluation protocol achieves an agreement of 79.9\% with human experts, indicating a strong ability of MLLM-as-a-Judge to simulate human preferences accurately. These findings demonstrate the viability of \name as an automatic, large-scale, and challenging benchmark for evaluating the assistance capabilities of MLLMs in professional productivity scenarios. By effectively aligning with human judgments, \name provides a reliable automatic framework for advancing MLLM development and assessment.

\paragraph{Future work and limitation.}  Although our \name has provided valuable insights into the performance and capabilities of MLLMs, several limitations remain that warrant further exploration. One key limitation is a potential bias in the benchmark tasks, which may not fully capture the diversity of real-world productivity scenarios for MLLMs. Future work could focus on expanding the benchmark to include a broader range of challenging tasks, potentially through the data synthesis (\eg, diffusion models and MLLMs), to improve the diversity. By addressing these challenges, \name can continue to evolve as a robust and comprehensive tool for advancing the development and evaluation of MLLMs.

\subsection{Distilled local evaluator} 
Considering the high API cost of using gpt-4o-2024-08-06 as the judge, we fine-tune a local evaluator to enable cost-effective and GPU-friendly evaluations for future MLLMs. We use the widely spread Llama-3.2-11B-Vision-Instruct as our backbone model. The Qwen and Pixtral MLLM families are reserved for testing, with the remaining data allocated for training. Our network is trained to distill both the reasoning and decisions of using gpt-4o-2024-08-06 as the judge. The network achieves an average root mean squared error of 32.58 in Elo ratings.

\section{Related work}
The evolution of MLLM-as-a-Judge is largely inspired by the concept of LLM-as-a-Judge \cite{li2024crowdsourced, dubois2024alpacafarm, zheng2023judging}, which aims to automatically measure the alignment between MLLMs and human preferences. While pairwise comparison \cite{li2024crowdsourced, chen2024mllm} is considered as most preferred, it suffers from biases introduced by factors such as the presentation order of MLLM outputs, verbosity, and markdown styles. To mitigate these issues, style control has been proposed \cite{li2024does}, using statistical modeling to de-bias these confounding effects, thereby improving the MLLM judges.

Other approaches, such as few-shot judging, have also been explored, but they face challenges such as reliance on the few-shot example selection and increased evaluation costs \cite{zheng2023judging}. Existing MLLM-as-a-Judge leaderboards can be specified to \cite{luo2024videoautoarena, lu2024wildvision, chen2024mllm}. However, these often focus on a narrow scope of MLLM capability dimensions \cite{luo2024videoautoarena, lu2024wildvision}, or rely on artificially posed evaluations by a limited number of human experts \cite{chen2024mega}, making them inadequate for assessing MLLMs on professional tasks. Consequently, they fail to capture the dynamic nature  of real-world human and MLLM interactions for a comprehensive assessment of MLLM capabilities. In contrast, this work introduces a challenging benchmark, \name,  curated from large-scale crowdsourced datasets reflecting real-world professional productivity scenarios. It features three distinct evaluation tracks: single-round, multi-round, and multi-linguistic conversations, 
across various task fields, offering a robust framework for evaluating MLLM performance in real-world
scenarios.

\section{Conclusion} 
This paper introduces the \name, which features single-round, multi-round, and multi-linguistic tracks to enable a comprehensive and challenging  assessment of the alignment between MLLMs and human preferences across diverse professional productivity demands. By employing MLLM-as-a-Judge, the benchmark evaluates MLLM pairwisely, achieving 79.9\% agreement with human expert judgments, and underscoring its reliability. Through benchmarking 24 leading MLLMs, our results reveal significant shortcomings of existing MLLMs, particularly in visual perception and reasoning. Furthermore, models often struggle with multi-linguistic and multi-round tracks, highlighting the challenges of diverse language requirement and complex interactions. It reveals valuable insights for future MLLM developments. We hope it inspires successors.

\bibliography{custom}

\clearpage
\appendix

\begin{table*}[!t]
    \centering
    \scriptsize
    \caption{\small Comparisons of state-of-the-art MLLMs on the multi-linguistic track are presented using the following abbreviations: PT (Portuguese), FR (French), ES (Spanish), DE (German), and an ``Other'' category (\eg, Chinese, Vietnamese, and more).  We provide ELO ratings for each language, followed by an overview that includes the average number of output tokens (\#Token), 95\% confidence interval (95\% CI), win rate (WR), and overall ELO rating. The MLLMs are sorted by the overall ELo rating in each group.
    }
    \vspace{-1em}
    \setlength{\tabcolsep}{3pt}
    \begin{tabularx}{\linewidth}{lr|YYYYY|cccc}
    \toprule
    \multirow{2}{*}{\bf Model} & & \multicolumn{5}{c|}{\bf Languge-Specific ELO Ratings} & \multicolumn{4}{c}{\bf Overview} \\
    \cmidrule{3-11}
    & & PT & FR &  ES & DE & Other &\#Token & 95\% CI & WR & Elo \\
    \multicolumn{2}{l|}{\textit{\textbf{\gc{Proprietary MLLMs}}}} & \multicolumn{5}{l|}{}\\ 
    \claudeemoji claude-3-5-sonnet-20241022    & \faLock{} & 1248  & 1319  & 1335  & 1389  & 1309  & 485 &  (-21, 29)   & 74.58 & 1301 \\
    \openaiemoji gpt-4o-2024-05-13 & \faLock{}  & 1114  & 1114  & 1114  & 1114  & 1114  & 585 &    (0, 0)    & 50.0 & 1114\\
    \googleemoji gemini-1.5-pro-002 & \faLock{} &  1273  & 1168  & 1131  & 1168  & 1139  & 629 &  (-20, 20)   & 59.11 & 1178 \\
    \openaiemoji gpt-4o-2024-08-06 & \faLock{}  &  1159  & 1224  & 1226  & 1259  & 1114  & 480 &  (-17, 26)   & 60.35 & 1187 \\
    \openaiemoji gpt-4o-mini-2024-07-18 & \faLock{} & 1038  & 1079  & 1071  & 1151  & 1099  & 657 &  (-21, 16)   & 45.84 & 1085 \\
    \googleemoji gemini-1.5-flash-002  & \faLock{} &  1031  &  990  &  845  & 1015  &  815  & 567 &  (-25, 19)   & 28.47 &  954 \\
    \midrule
    \multicolumn{2}{l|}{\textit{\textbf{\gc{70B+ Open-source MLLMs}}}} & \multicolumn{5}{l|}{} \\
    \pixtralemoji Pixtral-Large-Instruct-2411 & 124B &  1229  & 1496  & 1216  & 1324  & 1286  & 966 &  (-23, 22)   & 73.81 & 1294 \\
    \qwenemoji Qwen2-VL-72B-Instruct & 72B &  1067  & 1199  &  944  & 1241  &  999  & 834 &  (-18, 21)   & 47.56 & 1097 \\
    \internvlemoji InternVL2\_5-78B & 78B &  948  & 1125  & 1035  & 1123  & 1084  & 841 &  (-14, 20)   & 42.71 & 1063 \\
    \nvlmemoji NVLM-D-72B & 72B & 900  &  863  &  850  &  898  &  918  & 907 &  (-17, 25)   & 21.99 &  894\\
    \llamaemoji Llama-3.2-90B-Vision-Instruct & 90B &  905  &  860  &  824  &  863  &  864  & 968 &  (-29, 21)   & 20.92 &  883 \\
    \molmoemoji Molmo-72B-0924 & 72B & 834  &  835  &  852  &  853  &  878  & 426 &  (-27, 19)   & 18.9 &  861\\
    \llavaemoji llava-onevision-qwen2-72b-ov & 72B &  782  &  810  &  609  &  800  &  729  & 534 &  (-27, 24)   & 11.95 &  767 \\
    \midrule
    \multicolumn{2}{l|}{\textit{\textbf{\gc{10B+ Open-source MLLMs}}}} & \multicolumn{5}{l|}{} \\
    \internvlemoji InternVL2\_5-38B & 38B  & 1038  & 1092  & 1070  & 1100  & 1044  & 868 &  (-20, 18)   & 43.98 & 1072 \\
    \pixtralemoji Pixtral-12B-2409 & 12B &  935  & 1096  &  998  & 1077  &  929  & 1199 &  (-14, 22)   & 35.73 & 1012 \\
    \ariaemoji Aria-Chat & 3.9/25.3B&  964  & 1042  &  983  & 1041  &  999  & 1014 &  (-23, 17)   & 35.33 & 1009 \\
    \internvlemoji InternVL2\_5-26B & 26B & 779  &  858  &  782  &  880  &  839  & 814 &  (-28, 19)   & 17.7 &  847 \\
    \llamaemoji Llama-3.2-11B-Vision-Instruct & 11B &  714  &  663  &  626  &  627  &  665  & 2027 &  (-29, 21)   & 8.4 &  699 \\
    \midrule
    \multicolumn{2}{l|}{\textit{\textbf{\gc{7B+ Open-source MLLMs}}}} & \multicolumn{5}{l|}{} \\
    \qwenemoji Qwen2-VL-7B-Instruct & 7B & 701  &  875  &  673  &  865  &  678  & 1216 &  (-24, 22)   & 12.25 &  772 \\
    \internvlemoji InternVL2\_5-8B & 8B &  760  &  776  &  765  &  821  &  602  & 1021 &  (-22, 20)   & 11.95 &  767 \\
    \minicpmemoji MiniCPM-V-2\_6 & 8B & 522  &  559  &  603  &  634  &  455  & 890 &  (-36, 35)   & 4.44 &  581 \\
    \molmoemoji Molmo-7B-D-0924 & 7B & 445  &  495  &  577  &  613  &  505  & 406 &  (-52, 33)   & 4.32 &  576 \\
    \llavaemoji llava-onevision-qwen2-7b-ov & 7B & 579  &  386  &  144  &  403  &  588  & 686 &  (-68, 37)   & 3.07 &  514 \\
    \molmoemoji Molmo-7B-O-0924 & 7B & 383  &  256  &  536  &  246  &  429  & 512 &  (-73, 51)   & 1.95 &  433 \\
    \bottomrule
    \end{tabularx}
    \label{tab:multilinguistic}
\end{table*}

\begin{table*}[!t]
    \centering
    \scriptsize
    \caption{\small Comparisons of state-of-the-art MLLMs on the multiround track are presented. We provide ELO ratings for rounds with lengths of 2, 3, 4, 5, and more than 6 (6+), followed by an overview that includes the average number of output tokens (\#Token), 95\% confidence interval (95\% CI), win rate (WR), and overall ELO rating. `N/A' indicates cases where the model did not apply, as it lost to gpt-4o-2024-05-13 across all samples. The MLLMs are sorted by the overal ELo rating in each group
    }
    \vspace{-1em}
    \setlength{\tabcolsep}{3pt}
    \begin{tabularx}{\linewidth}{lr|YYYYY|YYYY}
    \toprule
    \multirow{2}{*}{\bf Model} & & \multicolumn{5}{c|}{\bf Round-Specific ELO Ratings} & \multicolumn{4}{c}{\bf Overview} \\
    \cmidrule{3-11}
    & & 2 & 3 &  4 & 5 & 6+ &\#Token & 95\% CI & WR & Elo \\
    \multicolumn{2}{l|}{\textit{\textbf{\gc{Proprietary MLLMs}}}} & \multicolumn{4}{l|}{}\\ 
    \claudeemoji claude-3-5-sonnet-20241022    & \faLock{} & 1260  & 1249  & 1356  & 1248  & 1321  & 1477 &  (-20, 18)   & 70.82 & 1268 \\
    \openaiemoji gpt-4o-2024-05-13 & \faLock{} & 1114  & 1114  & 1114  & 1114  & 1114  & 1563 &    (0, 0)    & 50.0 & 1114 \\
    \googleemoji gemini-1.5-pro-002 & \faLock{} & 1136  & 1140  & 1107  & 1207  & 1145  & 1425 &  (-26, 19)   & 53.88 & 1141 \\
    \openaiemoji gpt-4o-2024-08-06 & \faLock{}  & 1146  & 1050  & 1138  & 1023  &  965  & 1052 &  (-22, 18)   & 45.41 & 1082 \\
    \openaiemoji gpt-4o-mini-2024-07-18 & \faLock{} & 1147  & 1143  & 1142  & 1200  & 1151  & 1749 &  (-17, 24)   & 55.16 & 1150 \\
    \googleemoji gemini-1.5-flash-002  & \faLock{} & 1015  & 1040  & 1015  & 1119  & 1006  & 1388 &  (-16, 19)   & 38.14 & 1030 \\
    \midrule
    \multicolumn{2}{l|}{\textit{\textbf{\gc{70B+ Open-source MLLMs}}}} & \multicolumn{5}{l|}{} \\
    \pixtralemoji Pixtral-Large-Instruct-2411 & 124B & 1233  & 1273  & 1304  & 1376  & 1253  & 2593 &  (-23, 19)   & 69.73 & 1259  \\
    \qwenemoji Qwen2-VL-72B-Instruct & 72B & 1023  &  972  & 1033  &  936  &  875  & 1608 &  (-21, 19)   & 32.24 &  985 \\
    \internvlemoji InternVL2\_5-78B & 78B &  1135  & 1040  & 1148  & 1015  &  992  & 2015 &  (-21, 20)   & 44.84 & 1078   \\
    \nvlmemoji NVLM-D-72B & 72B &  770  &  557  &  602  &  641  &  682  & 1371 &  (-35, 33)   & 8.49 &  701 \\
    \llamaemoji Llama-3.2-90B-Vision-Instruct & 90B &  754  &  757  &  784  &  426  &  605  & 1350 &  (-36, 24)   & 9.88 &  730 \\
    \molmoemoji Molmo-72B-0924 & 72B & 886  &  817  &  787  &  920  &  808  & 967 &  (-28, 25)   & 18.64 &  858 \\
    \llavaemoji llava-onevision-qwen2-72b-ov & 72B &  753  &  721  &  673  &  525  &  692  & 1176 &  (-31, 26)   & 10.3 &  738 \\
    \midrule
    \multicolumn{2}{l|}{\textit{\textbf{\gc{10B+ Open-source MLLMs}}}} & \multicolumn{5}{l|}{} \\
    \internvlemoji InternVL2\_5-38B & 38B  & 1003  & 1037  & 1036  &  913  &  902  & 1734 &  (-18, 21)   & 34.68 & 1004 \\
    \pixtralemoji Pixtral-12B-2409 & 12B & 1054  & 1008  & 1160  & 1013  & 1035  & 2264 &  (-19, 20)   & 40.48 & 1047 \\
    \ariaemoji Aria-Chat & 3.9/25.3B&  937  &  913  &  946  &  887  &  812  & 2321 &  (-27, 12)   & 23.92 &  913 \\
    \internvlemoji InternVL2\_5-26B & 26B & 881  &  811  &  805  &  753  &  638  & 1554 &  (-27, 28)   & 15.77 &  823 \\
    \llamaemoji Llama-3.2-11B-Vision-Instruct & 11B & 741  &  380  &  487  &  275  &  490  & 2094 &  (-38, 32)   & 6.03 &  637 \\
    \midrule
    \multicolumn{2}{l|}{\textit{\textbf{\gc{7B+ Open-source MLLMs}}}} & \multicolumn{5}{l|}{} \\
    \qwenemoji Qwen2-VL-7B-Instruct & 7B & 808  &  622  &  637  &  557  &  495  & 2004 &  (-34, 25)   & 9.48 &  722 \\
    \internvlemoji InternVL2\_5-8B & 8B &  814  &  724  &  775  &  686  &  559  & 1835 &  (-25, 22)   & 11.77 &  764 \\
    \minicpmemoji MiniCPM-V-2\_6 & 8B &  664  &  575  &  628  &  530  &  389  & 1861 &  (-33, 37)   & 5.35 &  615 \\
    \molmoemoji Molmo-7B-D-0924 & 7B &  672  &  470  &  523  &  409  &  618  & 923 &  (-34, 26)   & 5.04 &  604 \\
    \llavaemoji llava-onevision-qwen2-7b-ov & 7B &  737  &  591  &  649  & N/A &  512  & 1743 &  (-30, 30)   & 6.58 &  653 \\
    \molmoemoji Molmo-7B-O-0924 & 7B & 589  &  413  &  490  & N/A &  402  & 925 &  (-49, 37)   & 3.43 &  534 \\
    \bottomrule
    \end{tabularx}
    \label{tab:multiround}
\end{table*}

\newcolumntype{P}[1]{>{\centering\arraybackslash}p{#1}}
\newcolumntype{Y}[1]{>{\centering\arraybackslash}m{#1}} 
\newcolumntype{L}[1]{>{\flushleft\arraybackslash}m{#1}}

\section{Experimental detail}
We respectively present detailed comparisons of multi-linguistic and multi-round tracks in \cref{tab:multilinguistic} and \cref{tab:multiround}.

The optimization details for tuning a local evaluator based on Llama-3.2-11B-Vision-Instruct are provided below. We use a learning rate of $1 \times 10^{-5}$ for both the projector and the LLM, while setting a lower learning rate of $2 \times 10^{-6}$ for the vision encoder. The context length is set to 128K. A cosine annealing strategy with a 3\% warm-up of the total optimization steps is employed. The AdamW optimizer is used with $\beta_{1}$ = 0.9 and $\beta_{1}$ = 0.95, along with a weight decay of $0.03$. We train with a batch size of 16 for 20K optimization steps. The model is trained using 16 H100 GPUs, with the training process taking approximately 2 days.

For evaluation with MLLM-as-the-Judge, the largest models require around two days for response generation on 8 GPUs, while evaluation with the local evaluator takes about one day using 2 GPUs. 

All data from ProBench has been collected with explicit user consent.

\section{Prompt template}
We present the prompts for curating the single-round, multi-linguistic, and multi-round tracks, as well as for utilizing MLLM-as-a-Judge across the three tracks:
\cref{table:querytype}, \cref{table:querysubtype}, \cref{table:query}, and \cref{table:image} provide prompts for categorizing task and sub-task fields related to user instruction queries;
\cref{table:chanllenge} and \cref{table:multiroundchanllenge} present prompts for evaluating challenges within user instruction queries;
\cref{table:deduplication} and \cref{table:multiroounddeduplication} are prompts for deduplications between visual and textual content in user instruction queries (\ie, image-instruction deduplication);
\cref{table:multirooundinterdepdency} offers prompts for assessing interdependencies among multi-round user instruction queries;
\cref{table:singleroundjudge}, \cref{table:multilinguisticjudge}, and \cref{table:multiroundjudge} respectively give the prompts of MLLM-as-a-Judge for the three tracks.

\section{Human preference evaluation}
To assess the agreements and reliability of MLLM-as-a-Judge, we evaluate the alignment between human annotators and gpt-4o-2024-08-06 as a judge. All participants are volunteers who have been informed about the purpose of the study and have provided consent to share their data. In this experiment, a random sample of 300 responses is drawn from the \name dataset. These responses are then evaluated by six human annotators, each tasked with comparing the outputs of two MLLMs for addressing the user instruction queries.

\begin{figure}[!t]
    \centering
    \includegraphics[width=\linewidth]{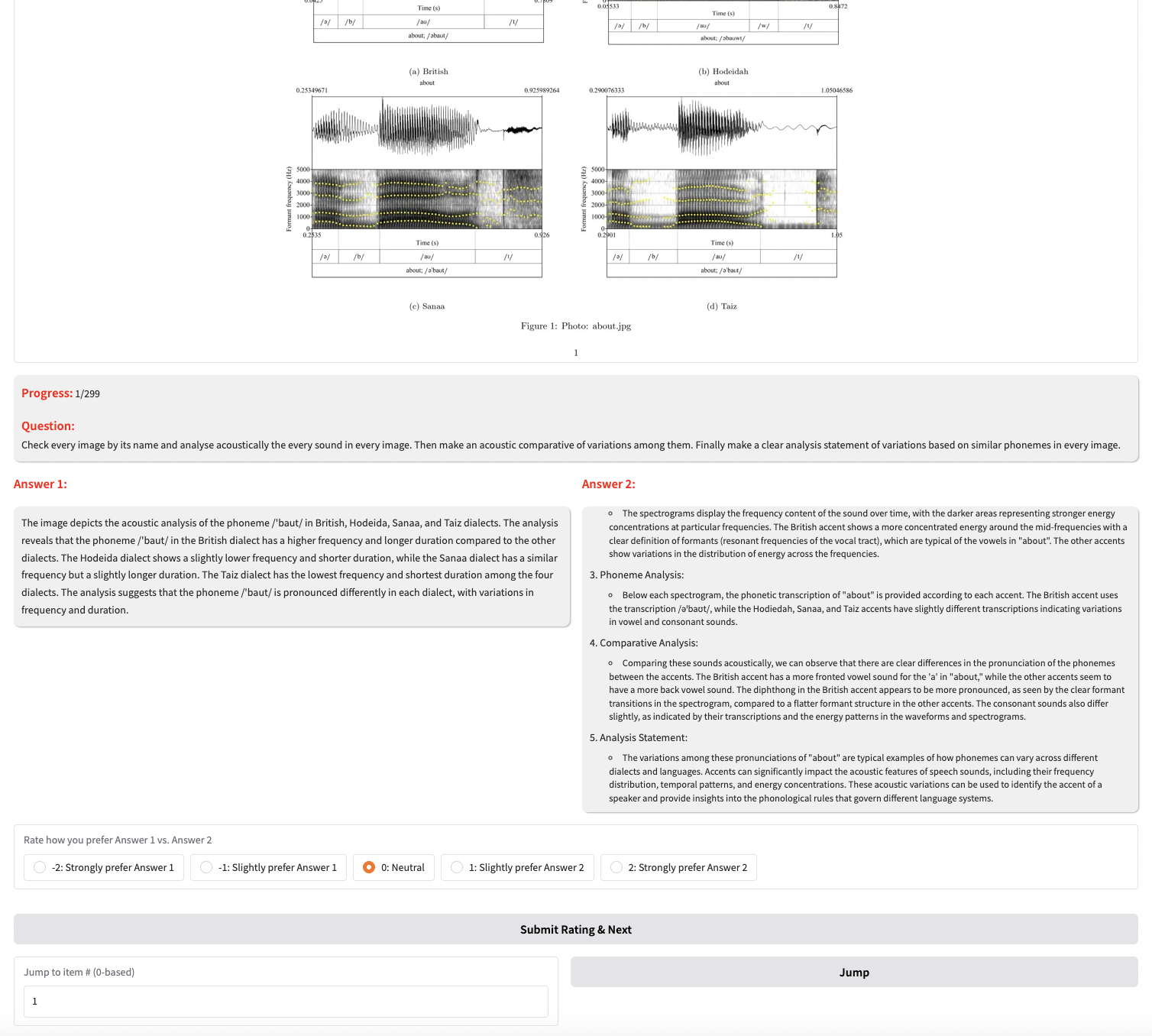}
    \caption{Interface used for human annotator to compare the outputs of multiple MLLMs.}
    \label{fig:comparemllm}
\end{figure}

\begin{figure*}[!t]
    \centering
    \includegraphics{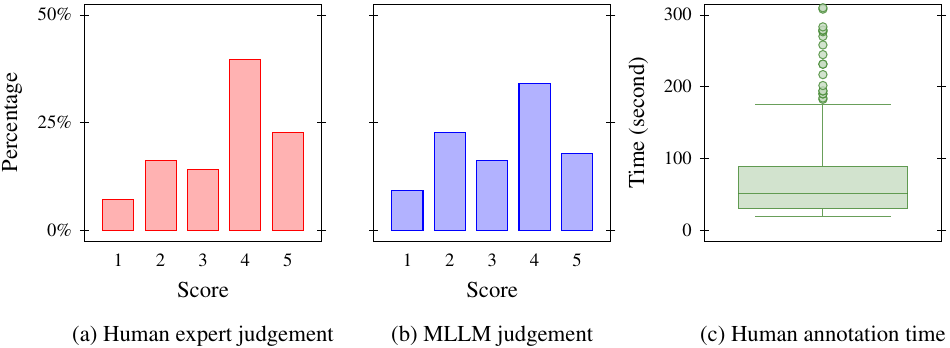} 
    \caption{Statistic of human expert annotation experiment. (a) Score distribution of human expert judgment. (b) Score distribution of MLLM-as-a-Judge. (c) Time distribution of human expert annotation.
    }
    \label{fig:distribution}
\end{figure*}

On average, each comparison took approximately 90.6 seconds. In contrast, the MLLM-as-a-Judge method completes the task in just a few seconds via an API call, highlighting the superior speed and efficiency of model-based evaluation. The annotation interface used for this task is shown in \cref{fig:comparemllm}. Overall, we observe 79.9\% agreement between human annotators and the MLLM-as-a-Judge. Refer to \cref{fig:distribution} that illustrates the distribution of human annotator preferences, MLLM preferences, and human annotation time cost.

\section{Analysis}
\begin{figure*}[!t]
    \centering
    \includegraphics{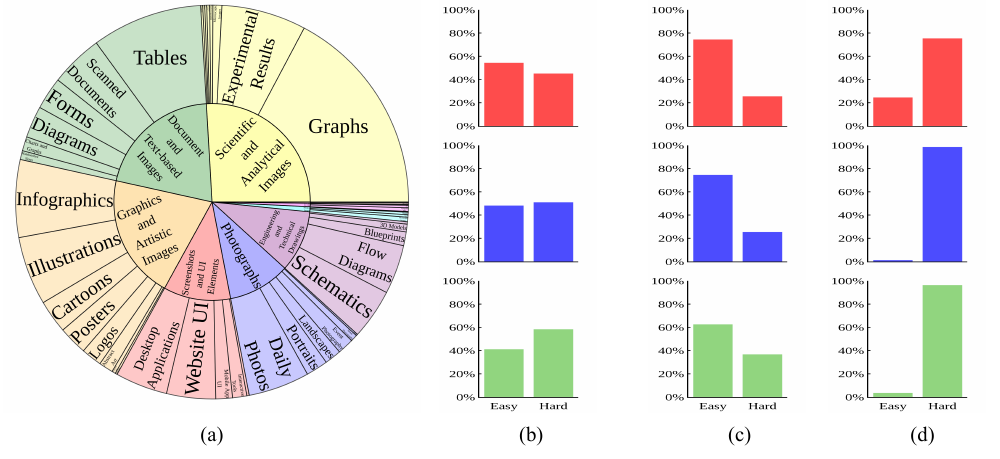}
    \caption{Distribution of (a) image fields, and challenge levels on (b) textual, (c) image, (d) reasoning parts of user queries across the three tracks. (b)-(c), we show the single-round, multi-linguistic, and multi-round tracks from top to bottom.}
    \label{fig:overviewsp}
\end{figure*}

In \cref{fig:overviewsp}, we further present the distributions of image distribution, textual challenges, image challenges, and reasoning challenges across the user instruction queries. \cref{table:compare} provides examples of MLLM-as-a-Judge evaluations, with key information highlighted in red to indicate correctness or errors.
%discriminative information highlighted in red.
%30 randomly sampled examples of MLLM-as-a-Judge evaluations.

\clearpage
\newpage
\onecolumn{
~
\captionof{table}{The prompt for identifying user instruction query task fields.} 
\vspace{-.5em}
\begin{tcolorbox}[
    colback=white, 
    colframe=black, 
    breakable, 
    enhanced jigsaw,
]
\colorbox{orange}{\textbf{[System]}}
\begin{lstlisting}
You are an AI assistant tasked with classifying a user-provided question and image into predefined categories. The question should be classified based on both the text of the question and the image provided, while the image classification should be based solely on the visual content of the image. Your responsibilities are:

1. Analyze the question and classify it under one category from the following list:
   - Coding: Focuses on code-related tasks such as debugging, generating, translating, and understanding programming logic.
   - Information Extraction: Involves tasks like extracting and analyzing details from data, structured parsing, summarization, and multimodal Q&A.
   - Knowledge: Covers arts, culture, fact-checking, and understanding diverse global and historical knowledge.
   - Mathematics: Includes problem-solving in algebra, calculus, geometry, number theory, graph theory, and numeric reasoning.
   - Metrics: Evaluates quality and performance in images, videos, papers, and other models or generated content.
   - Perception: Encompasses tasks like 3D understanding, image segmentation, multimodal captioning, and object or scene understanding.
   - Planning: Deals with creating strategies for agents, solving puzzles, reordering tasks, and planning complex processes.
   - Science: Applies to specialized domains like chemistry, physics, life sciences, and STEM-related problem-solving.
   - Creative Writing: Covers character development, storytelling, poetry, dialogue, scriptwriting, and worldbuilding across genres.
   - Arts and Humanities: Involves creative and cultural exploration, metaphorical thinking, narrative techniques, and genre-specific expression.

2. Classify the image into one of the main categories:
   - Document and Text-based Images: Includes scanned documents, forms, tables, and charts, used for record-keeping, data presentation, or analysis.
   - Medical Images: Diagnostic visuals like MRIs, X-rays, and pathology slides, used in healthcare and medical research.
   - Photographs: Everyday pictures, portraits, and landscapes captured with cameras, often for personal or professional use.
   - Scientific and Analytical Images: Specialized visuals like microscopic, astronomical, or spectrogram images for research and technical analysis.
   - Graphics and Artistic Images: Includes infographics, logos, cartoons, and illustrations for creative, branding, or informative purposes.
   - Screenshots and UI Elements: Captures of websites, apps, or software interfaces for documentation or demonstration.
   - Remote Sensing and Satellite Images: Aerial and satellite photos for mapping, monitoring, or geographic analysis.
   - Security and Surveillance: CCTV footage and thermal imaging for safety, monitoring, or investigative purposes.
   - Engineering and Technical Drawings: CAD designs, blueprints, and 3D models for architectural or engineering applications.
   - Specialized Formats: Includes barcodes, QR codes, fingerprints, and AR/VR visuals for unique or advanced use cases.

3. If the question or image does not fit existing categories, propose a new category with justification.

4. Do not generate the answer for the user question.

Your response should be in JSON format:
{
  "thinking_image": "Reasoning for your classification of image.",
  "image_category": "The category of the image."
  "thinking_question": "Reasoning for your classification of question.",
  "question_category": "The category of the user question.",
}
\end{lstlisting}

\label{table:querytype}
\end{tcolorbox}
}

{
\captionof{table}{The prompt for identifying user instruction query sub-task fields.} 
\vspace{-.5em}
\begin{tcolorbox}[
    colback=white, 
    colframe=black, 
    breakable,
    enhanced jigsaw,
]
\colorbox{orange}{\textbf{[System]}}
\begin{lstlisting}
    You are an AI assistant tasked with further classifying a user-provided question and image into sub-categories. The question should be classified based on both the text of the question and the image provided, while the image classification should be based solely on the visual content of the image. Your responsibilities are:
    
    1. **Question Classification**:
    
    - Analyze the question and assign it to the most relevant sub-category based on its content.
    - The question belongs to the main category "{question_category}" and should be classified into one of the following sub-categories:
        {question_subcats_formatted}

    2. **Image Classification**:
    
    - Analyze the image and assign it to the most relevant sub-category based solely on its visual content.
    - The image belongs to the main category "{image_category}" and should be classified into one of the following sub-categories:
        {image_subcats_formatted}

    3. If the question or image does not fit any of the above sub-categories, propose a new sub-category and provide a justification.
    
    4. Do not generate the answer for the user question.


    Your response must be structured in the following JSON format:
    {{
    "thinking_image": "Reasoning for the image sub-category classification.",
    "image_subcategory": "The sub-category for the image."
    "thinking_question": "Reasoning for the question sub-category classification.",
    "question_subcategory": "The sub-category for the question.",
    }}
\end{lstlisting}
\label{table:querysubtype}
\end{tcolorbox}
}

\newpage
{\captionof{table}{{The task and sub-task fields for user instruction queries (\eg, questions). For consistency, the naming convention aligns with \cref{table:querysubtype}. \texttt{question\_category} represents the task field, while \texttt{question\_subcats\_formatted} denotes the task sub-field.} \label{table:query}
\vspace{-.5em}
 }
\begin{longtable}{Y{.2\textwidth}l}
    \toprule
    \texttt{question\_category} & \texttt{question\_subcats\_formatted}  \\ \midrule
    \endfirsthead

    \midrule
    \texttt{question\_category} & \texttt{question\_subcats\_formatted} \\ \midrule
    \endhead

    \multicolumn{2}{r}{\textit{Continued on next page...}} \\ \midrule
    \endfoot

    \endlastfoot

\texttt{Information Extraction}& \begin{minipage}{0.8\textwidth} \begin{lstlisting} * Data Parsing: Extracting and organizing data from structured or semi-structured formats such as tables, logs, or files. 
* App Function Understanding: Analyzing and interpreting the purpose, features, and functionality of an application.
* Summarization: Condensing detailed information into a concise form while preserving key points and context.
* Entity Recognition: Identifying and categorizing specific elements such as names, dates, locations, or organizations.
* Relationship Mapping: Identifying and visualizing the connections or associations between different entities.
* Contextual Analysis: Understanding the meaning, intent, or relevance of data within its specific context.
\end{lstlisting} \end{minipage}\\ \midrule
\texttt{Creative Writing}& \begin{minipage}{0.8\textwidth} \begin{lstlisting} * Poetry: Crafting expressive, rhythmic, and often symbolic writing to convey emotions or ideas. 
* Storytelling: Developing compelling and engaging narratives for readers or audiences.
* Scriptwriting: Creating scripts for various media formats, including films, television, and plays.
* Worldbuilding: Designing intricate and immersive fictional settings, universes, or environments.
* Character Development: Creating, evolving, and deepening the personalities and arcs of fictional characters.
* Plot Structuring: Organizing the sequence of events and narrative flow to build tension, conflict, and resolution.
\end{lstlisting} \end{minipage}\\ \midrule
\texttt{Science}& \begin{minipage}{0.8\textwidth} \begin{lstlisting} * Chemistry: The study of matter, its composition, properties, and interactions at atomic and molecular levels. 
* Physics: The exploration of forces, motion, energy, and the fundamental nature of the universe.
* Biology: The study of living organisms, their functions, and interactions within ecosystems.
* Astronomy: The observation and study of celestial objects, space, and the physical universe as a whole.
* Life Science/Medical: The study of biological and medical sciences, including anatomy, physiology, and healthcare-related topics.
* STEM Problem-Solving: Using interdisciplinary approaches to tackle technical and scientific challenges.
\end{lstlisting} \end{minipage}\\ \midrule
\texttt{Knowledge}& \begin{minipage}{0.8\textwidth} \begin{lstlisting} * Historical Facts: Key details, events, and timelines from history, including their significance. 
* Human and Culture: Insights into human behavior, societal structures, traditions, and cultural practices.
* Scientific Knowledge: Understanding and explaining scientific concepts, theories, and principles across disciplines.
* World Knowledge: General information about global geography, politics, economies, and cultures.
* Fact-Checking: Verifying the accuracy of information and identifying misinformation or inaccuracies.
* Philosophical Inquiry: Exploring existential, ethical, and metaphysical questions to gain deeper understanding.
\end{lstlisting} \end{minipage}\\ \midrule
\texttt{Metrics}& \begin{minipage}{0.8\textwidth} \begin{lstlisting} * Image Quality: Evaluating the clarity, sharpness, and overall fidelity of visual content. 
* Model Performance: Assessing the accuracy, efficiency, and reliability of algorithms or machine learning models.
* Paper Review: Critiquing and analyzing research papers for quality, relevance, and scientific rigor.
* Content Evaluation: Judging the quality, coherence, and relevance of generated or provided content.
* Quality Assessment: Measuring and determining the overall standard or quality of various outputs or systems.
* Reward Models: Designing and evaluating models that provide feedback or incentives for optimizing performance in systems.
\end{lstlisting} \end{minipage}\\ \midrule
\texttt{Coding}& \begin{minipage}{0.8\textwidth} \begin{lstlisting} * Debugging: Identifying and fixing errors or issues in code. 
* Code Generation: Creating new code based on given requirements, templates, or problem-solving scenarios.
* Code Translation: Converting code from one programming language or framework to another.
* Code Optimization: Enhancing the efficiency, readability, and performance of existing code.
* Code Understanding: Interpreting and explaining the purpose, logic, or functionality of code.
\end{lstlisting} \end{minipage}\\ \midrule
\texttt{Perception}& \begin{minipage}{0.8\textwidth} \begin{lstlisting} * 3D Understanding: Analyzing and interpreting spatial structures, shapes, and three-dimensional relationships. 
* Counting: Identifying and quantifying the number of objects or elements in an image or scene.
* Multimodal Captioning: Generating descriptive captions by combining visual and textual data for an enriched understanding.
* Object Understanding: Recognizing, categorizing, and interpreting the attributes and roles of objects in visual content.
* Scene Understanding: Comprehending the arrangement, context, and interactions within a visual scene.
* Diagram and Document Understanding: Interpreting and extracting information from diagrams, charts, or text-based documents.
\end{lstlisting} \end{minipage}\\ \midrule
\texttt{Arts and Humanities}& \begin{minipage}{0.8\textwidth} \begin{lstlisting} * Creative Arts: Engaging in visual or performing arts. 
* Cultural Analysis: Examining societal norms and values.
* Narrative Techniques: Exploring storytelling methods.
* Genre-Specific Writing: Crafting work within specific literary or artistic genres.
\end{lstlisting} \end{minipage}\\ \midrule
\texttt{Mathematics}& \begin{minipage}{0.8\textwidth} \begin{lstlisting} * Algebra: Solving equations, understanding mathematical symbols, and exploring abstract structures. 
* Calculus: Analyzing rates of change and accumulation using derivatives and integrals.
* Function: Studying relationships between inputs and outputs, represented mathematically.
* Geometry: Exploring shapes, sizes, dimensions, and the properties of space.
* Graph Theory: Analyzing the relationships between nodes and edges in a network or graph.
* Number Theory: Investigating the properties, patterns, and relationships of numbers, especially integers.
* Statistics/Numerical Reasoning: Interpreting, analyzing, and presenting data to draw logical inferences and conclusions.
\end{lstlisting} \end{minipage}\\ \midrule
\texttt{Planning}& \begin{minipage}{0.8\textwidth} \begin{lstlisting} * Agent Strategy: Designing strategies and action plans for autonomous agents to achieve specific objectives. 
* Reordering: Resequencing tasks or events to optimize efficiency and effectiveness.
* Puzzle Solving: Finding logical or creative solutions to abstract, conceptual, or practical challenges.
* Game Strategy: Developing tactics, plans, and approaches to achieve success in game environments.
* Complex Workflow Design: Designing and managing intricate, multi-step processes to accomplish complex tasks or objectives.
\end{lstlisting} \end{minipage}\\ \midrule
\texttt{Other} & \texttt{Unspecified or generic category.} \\ \bottomrule
\end{longtable}}

{
\setcounter{table}{3}
\captionof{table}{{The field and sub-field for images in user instruction queries. For consistency, the naming convention aligns with \cref{table:querysubtype}. \texttt{image\_category} represents the image field, while \texttt{image\_subcats\_formatted} denotes the image sub-field.} \label{table:image}
\vspace{-.5em}
 }
\begin{longtable}{Y{.2\textwidth}l}
    \toprule
    \texttt{image\_category} & \texttt{image\_subcats\_formatted}  \\ \midrule
    \endfirsthead

    \midrule
    \texttt{image\_category} & \texttt{image\_subcats\_formatted} \\ \midrule
    \endhead

    \multicolumn{2}{r}{\textit{Continued on next page...}} \\ \midrule
    \endfoot

    \endlastfoot

\texttt{Screenshots and UI Elements}& \begin{minipage}{0.8\textwidth} \begin{lstlisting} * Website UI: Interfaces of websites. 
* Mobile App UI: User interfaces for mobile applications.
* Desktop Applications: Screenshots of software interfaces.
* Game Interfaces: Displays from video games.
* Interactive Tools: Screenshots of tools requiring user input.
\end{lstlisting} \end{minipage}\\ \midrule
\texttt{Document and Text-based Images}& \begin{minipage}{0.8\textwidth} \begin{lstlisting} * Forms: Structured documents designed to collect and organize specific information, such as surveys or applications. 
* Tables: Data systematically organized in rows and columns for easy analysis and interpretation.
* Scanned Documents: Digital copies of physical documents, often used for record-keeping or archival purposes.
* Charts and Graphs: Visual tools to represent data trends, comparisons, or distributions, such as bar charts, pie charts, or line graphs.
* Handwritten Notes: Freehand textual or graphical information, often informal or personal in nature.
* Diagrams: Illustrations that depict relationships, processes, systems, or concepts using symbols, shapes, and connections, such as flowcharts, mind maps, or organizational charts.
\end{lstlisting} \end{minipage}\\ \midrule
\texttt{Scientific and Analytical Images}& \begin{minipage}{0.8\textwidth} \begin{lstlisting} * Microscopic Images: High-magnification views of microscopic subjects. 
* Astronomical Images: Visuals of celestial objects or phenomena.
* Spectrograms: Graphs displaying signal frequencies over time.
* Graphs: Plots representing relationships between variables.
* Experimental Results: Visual data from scientific experiments.
\end{lstlisting} \end{minipage}\\ \midrule
\texttt{Engineering and Technical Drawings}& \begin{minipage}{0.8\textwidth} \begin{lstlisting} * CAD Designs: Computer-aided design blueprints. 
* Blueprints: Detailed architectural or engineering drawings.
* 3D Models: Digital representations of three-dimensional objects.
* Schematics: Diagrams showing systems or circuits.
* Flow Diagrams: Graphs representing processes or workflows.
\end{lstlisting} \end{minipage}\\ \midrule
\texttt{Medical Images}& \begin{minipage}{0.8\textwidth} \begin{lstlisting} * X-rays: Radiographic images used to visualize internal structures of the body, such as bones and lungs. 
* MRIs: High-resolution imaging using magnetic resonance technology to capture detailed views of organs and tissues.
* Pathology Slides: Microscopic images of tissues or cells used for diagnosing diseases.
* Ultrasound: Images produced using sound waves to visualize internal body structures, commonly used in prenatal and organ assessments.
* Microscopic Images: Magnified visuals of biological specimens, such as cells or microorganisms, for medical analysis.
* CT Scans: Cross-sectional images of the body generated using computed tomography to provide detailed anatomical views.
\end{lstlisting} \end{minipage}\\ \midrule
\texttt{Photographs}& \begin{minipage}{0.8\textwidth} \begin{lstlisting} * Portraits: Photographs focusing on individuals or groups, often emphasizing expression and personality. 
* Landscapes: Scenic views showcasing natural environments or urban settings, often highlighting beauty or scale.
* Wildlife: Images capturing animals in their natural habitats, emphasizing behavior and environment.
* Street Photography: Candid shots portraying urban life, capturing everyday moments and street scenes.
* Event Photography: Documenting significant occasions such as weddings, conferences, or celebrations.
* Daily Photos: Casual and informal photographs capturing everyday moments, activities, or surroundings.
\end{lstlisting} \end{minipage}\\ \midrule
\texttt{Graphics and Artistic Images}& \begin{minipage}{0.8\textwidth} \begin{lstlisting} * Infographics: Visual representations of data or information designed to communicate complex ideas clearly and efficiently. 
* Logos: Graphic symbols or emblems used to identify brands, companies, or organizations.
* Cartoons: Illustrations with a humorous, exaggerated, or narrative style, often used in storytelling or entertainment.
* Illustrations: Artistic visuals created to complement text or communicate creative ideas.
* Posters: Artistic layouts designed for advertisements, events, or promotions.
* Abstract Art: Creative visuals emphasizing color, shape, and form without specific subjects.
* Typography Art: Designs focusing on stylized text and fonts to create visual impact.
\end{lstlisting} \end{minipage}\\ \midrule
\texttt{Remote Sensing and Satellite Images}& \begin{minipage}{0.8\textwidth} \begin{lstlisting} * Aerial Photos: Images taken from aircraft. 
* Thermal Images: Heat-map visuals for temperature analysis.
* Multispectral Images: Images across various light wavelengths.
* Topographic Maps: Maps showing elevation and terrain features.
\end{lstlisting} \end{minipage}\\ \midrule
\texttt{Specialized Formats}& \begin{minipage}{0.8\textwidth} \begin{lstlisting} * Barcodes: Visual codes representing data. 
* QR Codes: Two-dimensional codes for quick scanning.
* Fingerprints: Unique ridged patterns for identification.
* AR/VR Visuals: Content designed for augmented or virtual reality.
\end{lstlisting} \end{minipage}\\ \midrule
\texttt{Other} & \texttt{Unspecified or generic category.} \\ \bottomrule
\end{longtable}}

{
\setcounter{table}{4}
\captionof{table}{The prompt for identifying user instruction challenge in the single-round  track and multi-linguistic track. Scores below 6 are considered easy, while scores of 6 or higher are classified as hard.} 
\vspace{-.5em}
\begin{tcolorbox}[
    colback=white, 
    colframe=black, 
    breakable,
    enhanced jigsaw,
]
\colorbox{orange}{\textbf{[System]}}
\begin{lstlisting}
You are an AI assistant tasked with assessing the challenges of answering a user-provided question that combines textual instructions and visual images. A reference answer will be provided to guide your assessment.

### Input Format:
The input consists of three components in the following order:
1. Visual Images: One or more images relevant to the question.
2. Textual Instruction: Enclosed in <inst/> tags.
3. Reference Answer: Enclosed in <answer/> tags.

{images}

Textual Instruction:
<inst/>
{instruction text}
</inst/>

Reference Answer:
<answer/>
{reference answer}
</answer/>

### Scoring Criteria
Evaluate the difficulty across three dimensions using a scale of 1-10, where higher scores indicate greater difficulty:

1. Textual Complexity (How complex is the instruction?):
    - (1.1) Score 0: The instruction is redundantly presented in both visual and textual content.
    - (1.2) Score 1-3: Simple, straightforward instructions with minimal requirements and no domain knowledge needed.
    - (1.3) Score 4-6: Moderately complex instructions with some context and basic domain knowledge required.
    - (1.4) Score 7-9: Complex instructions with multiple requirements and specialized domain knowledge needed.
    - (1.5) Score 10: Highly complex instructions requiring significant expertise and precise understanding.

2. Visual Complexity (How complex are the images?)
    - (2.1) Score 0: The visual content merely duplicates the textual instruction.
    - (2.2) Score 1-3: Simple images with clear, distinct elements requiring minimal interpretation.
    - (2.3) Score 4-6: Moderately complex images with multiple elements requiring basic interpretation.
    - (2.4) Score 7-9: Complex images with multiple interrelated elements requiring domain knowledge.
    - (2.5) Score 10: Highly complex images requiring specialized expertise to interpret.

3. Reasoning Complexity (How complex is the integration of text and image?)
    - (3.1) Score 0: Question can be answered using text alone, images are unnecessary.
    - (3.2) Score 1-3: Simple reasoning requiring basic observation of text and images.
    - (3.3) Score 4-6: Moderate reasoning requiring integration of text and images with basic domain knowledge.
    - (3.4) Score 7-9: Complex reasoning requiring careful integration of text and images with specialized knowledge.
    - (3.5) Score 10: Advanced multi-step reasoning requiring expert knowledge to integrate complex text and images.


### Important Notes:
- Focus only on difficulty assessment - do not attempt to answer the question.
- Provide specific examples from the input when explaining scores.
- Consider the reference answer's approach when evaluating complexity.
- Each dimension must be scored independently.

### Response Format:
Provide your assessment in the following JSON structure:
{
    "challenge_textual": {
        "explanation": "Detailed explanation referencing specific scoring criteria (1.1-1.5) and examples from the input",
        "score": Integer value between 0-10
    },
    "challenge_image": {
        "explanation": "Detailed explanation referencing specific scoring criteria (2.1-2.5) and examples from the input",
        "score": Integer value between 0-10
    },
    "challenge_reasoning": {
        "explanation": "Detailed explanation referencing specific scoring criteria (3.1-3.5) and examples from the input",
        "score": Integer value between 0-10
    }
}
\end{lstlisting}
\label{table:chanllenge}
\end{tcolorbox}
}

{
\captionof{table}{The prompt for identifying user instruction challenge in the multi-round track. Scores below 6 are considered easy, while scores of 6 or higher are classified as hard.} 
\vspace{-.5em}
\begin{tcolorbox}[
    colback=white, 
    colframe=black, 
    breakable,
    enhanced jigsaw,
]
\colorbox{orange}{\textbf{[System]}}
\begin{lstlisting}
You are an AI assistant tasked with assessing the challenges of answering a user-provided question that combines textual instructions and visual images. A reference answer will be provided to guide your assessment.

### Input Format:
The input consists of two primary components:
1. Visual Images: One or more images relevant to the question.
2. Each turn which is Enclosed by <turn{number}> contains:
    - Textual Instruction: Enclosed in <inst/> tags
    - Reference Answer: Enclosed in <ans/> tags

{images}

<turn{number}/>
Textual Instruction:
<inst/>
{instruction text}
</inst>

Reference Answer:
<ans/>
{reference answer}
</ans>
</turn{number}>


### Scoring Criteria
Evaluate the difficulty across three dimensions using a scale of 1-10, where higher scores indicate greater difficulty:

1. Textual Complexity (How complex is the instruction?):
    - (1.1) Score 0: The instruction is redundantly presented in both visual and textual content.
    - (1.2) Score 1-3: Simple, straightforward instructions with minimal requirements and no domain knowledge needed.
    - (1.3) Score 4-6: Moderately complex instructions with some context and basic domain knowledge required.
    - (1.4) Score 7-9: Complex instructions with multiple requirements and specialized domain knowledge needed.
    - (1.5) Score 10: Highly complex instructions requiring significant expertise and precise understanding.

2. Visual Complexity (How complex are the images?)
    - (2.1) Score 0: The visual content merely duplicates the textual instruction.
    - (2.2) Score 1-3: Simple images with clear, distinct elements requiring minimal interpretation.
    - (2.3) Score 4-6: Moderately complex images with multiple elements requiring basic interpretation.
    - (2.4) Score 7-9: Complex images with multiple interrelated elements requiring domain knowledge.
    - (2.5) Score 10: Highly complex images requiring specialized expertise to interpret.

3. Reasoning Complexity (How complex is the integration of text and image?)
    - (3.1) Score 0: Question can be answered using text alone, images are unnecessary.
    - (3.2) Score 1-3: Simple reasoning requiring basic observation of text and images.
    - (3.3) Score 4-6: Moderate reasoning requiring integration of text and images with basic domain knowledge.
    - (3.4) Score 7-9: Complex reasoning requiring careful integration of text and images with specialized knowledge.
    - (3.5) Score 10: Advanced multi-step reasoning requiring expert knowledge to integrate complex text and images.


### Important Notes:
- Focus only on difficulty assessment - do not attempt to answer the question.
- Provide specific examples from the input when explaining scores.
- Consider the reference answer's approach when evaluating complexity.
- Each dimension must be scored independently.

### Response Format:
Provide your assessment in the following JSON structure:
{
    "challenge_textual": {
        "explanation": "Detailed explanation referencing specific scoring criteria (1.1-1.5) and examples from the input",
        "score": Integer value between 0-10
    },
    "challenge_image": {
        "explanation": "Detailed explanation referencing specific scoring criteria (2.1-2.5) and examples from the input",
        "score": Integer value between 0-10
    },
    "challenge_reasoning": {
        "explanation": "Detailed explanation referencing specific scoring criteria (3.1-3.5) and examples from the input",
        "score": Integer value between 0-10
    }
}
\end{lstlisting}
\label{table:multiroundchanllenge}
\end{tcolorbox}
}

{
\captionof{table}{The prompt for image-instruction deduplication in the single-round track and multi-linguistic track.} 
\vspace{-.5em}
\begin{tcolorbox}[
    colback=white, 
    colframe=black, 
    breakable,
    enhanced jigsaw,
]
\colorbox{orange}{\textbf{[System]}}
\begin{lstlisting}
You are an AI assistant tasked with determining whether a user question can be answered solely by the textual instruction, when a user provides both visual images and a textual instruction.

### Input Format:
The input consists of two primary components:
1. Visual Images: One or more images relevant to the question
2. Textual Instruction: Enclosed in <inst/> tags

{images}

Textual Instruction:
<inst/>
{instruction text}
<inst/>

### Evaluation Criteria:
- Carefully analyze the textual instruction and the associated question.
- Assess whether the ENTIRE question can be comprehensively answered using ONLY the text provided.

### Decision Guidelines:
- YES: If the textual instruction provides comprehensive, unambiguous information to answer the question
- NO: If any critical piece of information is missing or requires visual interpretation to answer the question


### Response Format:
Provide your assessment in the following JSON structure:
{
    "reasoning": "Clearly outline your analysis and explain the logic behind your conclusion.",
    "decision": "YES or NO"
}
\end{lstlisting}
\label{table:deduplication}
\end{tcolorbox}
}

{
\captionof{table}{The prompt for image-instruction deduplication in the multi-round track.} 
\vspace{-.5em}
\begin{tcolorbox}[
    colback=white, 
    colframe=black, 
    breakable,
    enhanced jigsaw,
]
\colorbox{orange}{\textbf{[System]}}
\begin{lstlisting}
You are an AI assistant tasked with evaluating the dependency of textual instructions on visual information across a multi-turn conversation.

### Input Format:
The input consists of two primary components:
1. Visual Images: Provided at the beginning of the conversation
2. Each turn which is Enclosed by <turn{number}> contains:
    - Textual Instruction: Enclosed in <inst/> tags
    - Answers: Enclosed in <ans/> tags

{images}

<turn{number}/>
Textual Instruction:
<inst/>
{instruction text}
<inst/>

Answers:
<ans/>
{answer text}
<ans/>
</turn{number}>

{More continuing conversation turns...}


### Evaluation Criteria:
- Carefully analyze the textual instruction from ALL conversation turns
- Assess whether the ENTIRE set of instructions can be comprehensively answered without using the visual/image information
- Consider the cumulative context and details from all turns.


### Decision Guidelines:
- YES: If textual instructions across all turns can be fully understood and addressed without relying on the visual/image information
- NO: If any critical piece of information is missing or requires visual interpretation to answer the question

### Response Format:
Provide your assessment in the following JSON structure:
{
    "reasoning": "Clearly outline your analysis and explain the logic behind your conclusion.",
    "decision": "YES or NO"
}
\end{lstlisting}
\label{table:multiroounddeduplication}
\end{tcolorbox}
}

{
\captionof{table}{The prompt for assessing interdependency among user instruction queries in the multi-round track.} 
\vspace{-.5em}
\begin{tcolorbox}[
    colback=white, 
    colframe=black, 
    breakable,
    enhanced jigsaw,
]
\colorbox{orange}{\textbf{[System]}}
\begin{lstlisting}
You are an AI assistant tasked with determining whether the turns in a multi-turn conversation are independent or interconnected.

### Input Format:
The input consists of two primary components:
1. Visual Images: Provided at the beginning of the conversation
2. Each turn which is Enclosed by <turn{number}> contains:
    - Textual Instruction: Enclosed in <inst/> tags
    - Answers: Enclosed in <ans/> tags

{images}

<turn{number}/>
Textual Instruction:
<inst/>
{instruction text}
<inst/>

Answers:
<ans/>
{answer text}
<ans/>
</turn{number}>

{More continuing conversation turns...}


### Independence Criteria:
Independent Turns: 
  - Each turn can be understood and are answered in isolation
  - No contextual dependency between turns
  - No clear progression or building upon previous turns

Interconnected Turns:
  - Turns have logical progression, i.e., later turns depend on context from earlier turns
  - Conversation follows a coherent narrative or problem-solving flow

### Decision Guidelines:
- YES: If turns are completely independent
- NO: If turns are interconnected and cannot be meaningfully separated


### Response Format:
Provide your assessment in the following JSON structure:
{
    "reasoning": "Clearly outline your analysis and explain the logic behind your conclusion.",
    "decision": "YES or NO"
}
\end{lstlisting}
\label{table:multirooundinterdepdency}
\end{tcolorbox}
}

\clearpage
{
\captionof{table}{The prompt for MLLM-as-a-Judge for the single-round track.} 
\vspace{-.5em}
\begin{tcolorbox}[
    colback=white, 
    colframe=black, 
    breakable,
    enhanced jigsaw,
]
\colorbox{orange}{\textbf{[System]}}
\begin{lstlisting}
You are an impartial judge tasked with evaluating two AI assistants' responses to a given prompt involving textual instructions and visual images.

### Evaluation Framework

#### Generate Your Own Answer
1. Generate an independent,  high-quality answer to the original prompt
2. Serves as a benchmark for comparison
3. Demonstrates the ideal response approach

#### Evaluation Dimensions
Assess the assistants' answers based on the following dimensions:

1. Correctness
   - Accuracy of information
   - Absence of factual and demonstrable errors
   - Alignment with known knowledge and visual evidence

2. Helpfulness
   - Directly addresses the user's instructions
   - Provides clear and practical guidance
   - Anticipates and resolves potential user questions

3. Relevance
   - Stringent focus on the prompt requirements
   - Eliminates extraneous or tangential information
   - Maintains precise topical alignment

4. Conciseness
   - Delivers information efficiently
   - Avoids unnecessary verbosity
   - Uses clear, direct language

5. Completeness
   - Covers all essential aspects of the prompt
   - Provides sufficient information to fully address the user's needs

#### Comparative Analysis
- Directly compare Assistant A and Assistant B's responses
- Nuanced evaluation of relative strengths and weaknesses
- Evidence-based assessment with specific textual references

#### Judgment Guidelines
1. Avoid any position biases and ensure that the order in which the assistants'responses were presented does not influence your decision
2. When the prompt contains ambiguity:
- Prioritize requesting clarification over making assumptions
- Evaluate how well each assistant handles potential uncertainties

### Input Format
1. Visual Images: Relevant images
2. Textual Instruction: Enclosed in <inst/> tags
3. Assistant A's Answer: Enclosed in <a/> tags
4. Assistant B's Answer: Enclosed in <b/> tags

{images}

Textual Instruction:
<inst/>
{instruction text}
</inst>

Assistant A's Answer:
<a/>
{Answers from Assistant A}
</a>

Assistant B's Answer:
<b/>
{Answers from Assistant B}
</b>

### Response Format

Answer: 
[Your comprehensive answer to the prompt]

Detailed Explanation:
[Thorough, point-by-point comparison of Assistant A and B's responses]

Specific Observations:
- Correctness assessment
- Helpfulness evaluation
- Relevance analysis
- Conciseness review
- Completeness check

Final Verdict:
Select ONE of the following:
- [[A>>B]]: Assistant A is significantly better
- [[A>B]]: Assistant A is slightly better
- [[A=B]]: Tie, relatively the same
- [[B>A]]: Assistant B is slightly better
- [[B>>A]]: Assistant B is significantly better
\end{lstlisting}
\label{table:singleroundjudge}
\end{tcolorbox}
}

{
\captionof{table}{The prompt for MLLM-as-a-Judge for the multi-linguistic track.} 
\vspace{-.5em}
\begin{tcolorbox}[
    colback=white, 
    colframe=black, 
    breakable,
    enhanced jigsaw,
]
\colorbox{orange}{\textbf{[System]}}
\begin{lstlisting}
You are an impartial judge tasked with evaluating two AI assistants' responses to a given prompt involving textual instructions and visual images.

### Evaluation Framework

#### Generate Your Own Answer
1. Generate an independent,  high-quality answer to the original prompt
2. Serves as a benchmark for comparison
3. Demonstrates the ideal response approach

#### Evaluation Dimensions
Assess the assistants' answers based on the following dimensions:

1. Correctness
   - Accuracy of information
   - Absence of factual and demonstrable errors
   - Alignment with known knowledge and visual evidence
   - Response must be in the same language as the textual instruction (unless explicitly specified otherwise)

2. Helpfulness
   - Directly addresses the user's instructions
   - Provides clear and practical guidance
   - Anticipates and resolves potential user questions

3. Relevance
   - Stringent focus on the prompt requirements
   - Eliminates extraneous or tangential information
   - Maintains precise topical alignment

4. Conciseness
   - Delivers information efficiently
   - Avoids unnecessary verbosity
   - Uses clear, direct language

5. Completeness
   - Covers all essential aspects of the prompt
   - Provides sufficient information to fully address the user's needs

#### Comparative Analysis
- Directly compare Assistant A and Assistant B's responses
- Nuanced evaluation of relative strengths and weaknesses
- Evidence-based assessment with specific textual references

#### Judgment Guidelines
1. Avoid any position biases and ensure that the order in which the assistants'responses were presented does not influence your decision
2. When the prompt contains ambiguity:
- Prioritize requesting clarification over making assumptions
- Evaluate how well each assistant handles potential uncertainties

### Input Format
1. Visual Images: Relevant images
2. Textual Instruction: Enclosed in <inst/> tags
3. Assistant A's Answer: Enclosed in <a/> tags
4. Assistant B's Answer: Enclosed in <b/> tags

{images}

Textual Instruction:
<inst/>
{instruction text}
</inst>

Assistant A's Answer:
<a/>
{Answers from Assistant A}
</a>

Assistant B's Answer:
<b/>
{Answers from Assistant B}
</b>

### Response Format

Answer: 
[Your comprehensive answer to the prompt]

Detailed Explanation:
[Thorough, point-by-point comparison of Assistant A and B's responses]

Specific Observations:
- Correctness assessment
- Helpfulness evaluation
- Relevance analysis
- Conciseness review
- Completeness check

Final Verdict:
Select ONE of the following:
- [[A>>B]]: Assistant A is significantly better
- [[A>B]]: Assistant A is slightly better
- [[A=B]]: Tie, relatively the same
- [[B>A]]: Assistant B is slightly better
- [[B>>A]]: Assistant B is significantly better
\end{lstlisting}
\label{table:multilinguisticjudge}
\end{tcolorbox}
}

{
\captionof{table}{The prompt for MLLM-as-a-Judge for the multi-round track.} 
\vspace{-.5em}
\begin{tcolorbox}[
    colback=white, 
    colframe=black, 
    breakable,
    enhanced jigsaw,
]
\colorbox{orange}{\textbf{[System]}}
\begin{lstlisting}
You are an impartial judge tasked with evaluating two AI assistants' responses to given prompts involving textual instructions and visual images.

### Evaluation Framework

#### Generate Your Own Answer
1. Generate an independent, high-quality answer to the original prompt
2. Serves as a benchmark for comparison
3. Demonstrates the ideal response approach

#### Evaluation Dimensions
Assess the assistants' answers based on the following dimensions:

1. Correctness
   - Accuracy of information
   - Absence of factual and demonstrable errors
   - Alignment with known knowledge and visual evidence

2. Helpfulness
   - Directly addresses the user's instructions
   - Provides clear and practical guidance
   - Anticipates and resolves potential user questions

3. Relevance
   - Stringent focus on the prompt requirements
   - Eliminates extraneous or tangential information
   - Maintains precise topical alignment

4. Conciseness
   - Delivers information efficiently
   - Avoids unnecessary verbosity
   - Uses clear, direct language

5. Completeness
   - Covers all essential aspects of the prompt
   - Provides sufficient information to fully address the user's needs

#### Comparative Analysis
- Directly compare Assistant A and Assistant B's responses
- Nuanced evaluation of relative strengths and weaknesses
- Evidence-based assessment with specific textual references

#### Judgment Guidelines
1. Avoid any position biases and ensure that the order in which the assistants'responses were presented does not influence your decision
2. When the prompt contains ambiguity:
- Prioritize requesting clarification over making assumptions
- Evaluate how well each assistant handles potential uncertainties

### Input Format
1. Visual Images: Relevant images
2. Textual Instruction: Enclosed in <inst/> tags
3. Turn Structure: Each turn is enclosed with <turn{number}> tags
4. Assistant A's Answer: Enclosed in <a/> tags
5. Assistant B's Answer: Enclosed in <b/> tags

{images}

<turn{number}/>
Textual Instruction:
<inst/>
{instruction text}
</inst>

Assistant A's Answer:
<a/>
{Answers from Assistant A}
</a>

Assistant B's Answer:
<b/>
{Answers from Assistant B}
</b>
</turn{number}>

### Response Format

Answer: 
[Your comprehensive answer to the prompt]

Detailed Explanation:
[Thorough, point-by-point comparison of Assistant A and B's responses]

Specific Observations:
- Correctness assessment
- Helpfulness evaluation
- Relevance analysis
- Conciseness review
- Completeness check

Final Verdict:
Select ONE of the following:
- [[A>>B]]: Assistant A is significantly better
- [[A>B]]: Assistant A is slightly better
- [[A=B]]: Tie, relatively the same
- [[B>A]]: Assistant B is slightly better
- [[B>>A]]: Assistant B is significantly better
\end{lstlisting}
\label{table:multiroundjudge}
\end{tcolorbox}
}

\newpage 

\clearpage
\newpage
{\captionof{table}{{Examples of MLLM-as-a-Judge with key information highlighted in red to indicate the correctness or errors.} \label{table:compare}
\vspace{-.5em}
a }

 \lstset{
      aboveskip=0pt, 
      belowskip=0pt,   
      basicstyle=\ttfamily,
      inputencoding=utf8,  
      breaklines=true,         
      breakatwhitespace=true,   
      breakindent=0pt,
      columns=flexible, 
      escapeinside={(*@}{@*)},
      literate={
        {’}{{\textquoteleft}}1 
        {–}{{\textendash}}1
        {—}{{\textemdash}}1 
        {‘}{{\textquoteleft}}1 
        {’}{{\textquoteright}}1 
        {“}{{\textquotedblleft}}1 
        {”}{{\textquotedblright}}1 
        {²}{{\textsuperscript{2}}}1 
        {✓}{{\checkmark}}1 
        {●}{{\textbullet}}1
        {○}{{\textcircled{1}}}1 
        {∈}{{$\in$}}1          
        {=}{{\texttt{=}}}1   
        {∪}{{$\cup$}}1           
        {∩}{{$\cap$}}1           
        {≤}{{$\leq$}}1           
        {≥}{{$\geq$}}1           
        {^}{{**}}1
        {μ}{{$\mu$}}1            
        {σ}{{$\sigma$}}1 
        {°}{{\textdegree}}1 
    }
    }

\begin{tcolorbox}[
    colback=white, 
    colframe=black, 
    breakable,
    enhanced jigsaw,
]

    \colorbox{orange}{\textbf{Image:}} 
    
    \includegraphics[width=\linewidth, height=.5\linewidth, valign=t]{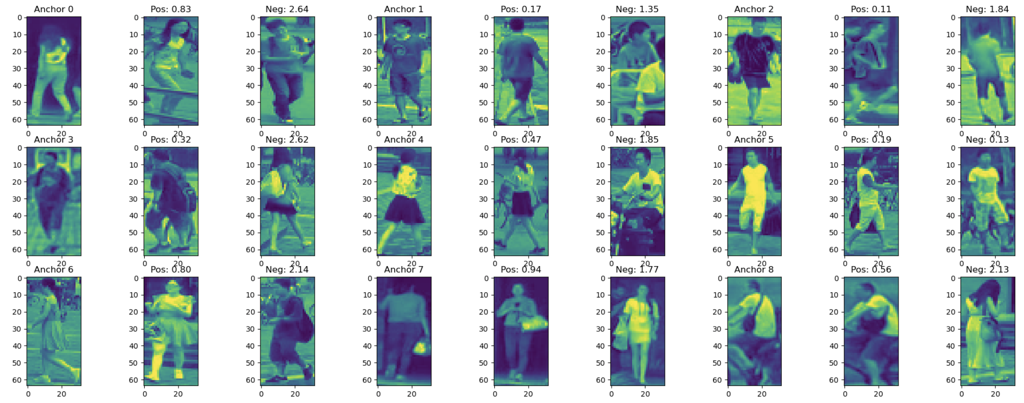}   
    
    \colorbox{yellow}{\textbf{Question:}}

    \begin{lstlisting}
what does this say about a triplet loss model for a person re identification task, is the model performing well or not? go into full detail and explain
\end{lstlisting}

    \colorbox{gray}{\textbf{gpt-4o-05-13:}} 
  
  \begin{lstlisting}
The image displays a set of results from a triplet loss model used for person re-identification tasks. Each column in the image represents a separate example, where three images are shown: one anchor image, a positive example (Pos), and a negative example (Neg). The numerical values next to the positive and negative labels are likely the computed distances or similarity scores from the anchor image.

### Analysis of Model Performance:

1. **Interpretation of Scores:**
   - **Positive (Pos) Score:** This score {measures the similarity or closeness between the anchor image and the positive image (images of the same person ideally). (*@\textcolor{red}{A lower score indicates higher similarity}@*).
   - **Negative (Neg) Score:** This score measures the similarity between the anchor image and the negative image (images of different persons). (*@\textcolor{red}{\mbox{A higher score indicates greater dissimilarity, which is desired}}@*).

2. **Expected Model Behavior:**
   - Ideally, for the model to perform well, (*@\textcolor{red}{the positive score should be significantly lower than the \mbox{negative score}}@*). This indicates that the model can effectively differentiate between images of the same person and images of different people.

3. **Observations from the Image:**
   - In most cases, the negative scores are higher than the positive scores, which suggests that the model can distinguish between different individuals to some extent. For example, (*@\textcolor{red}{in Anchor 0, the positive score is 0.83, and the negative score is 2.64, which is a good separation}@*).
   - However, some examples show closer scores or inappropriate scoring. For example, in Anchor 1, the positive score is 0.17, and the negative score is 1.35. Although the negative is still higher, the difference is not as pronounced, which might indicate potential issues in cases with subtle differences or challenging scenarios.

4. **Performance Consistency:**
   - The performance seems varied across different examples. This variation could depend on factors like the pose of the person, the lighting conditions, and the background. Models trained on more diverse datasets tend to handle such variations better.

5. **Further Evaluation:**
   - Quantitative metrics such as accuracy, precision, recall, and F1-score on a labeled test set would provide more concrete evidence of model performance.
   - Analyzing cases where the model fails or performs poorly could help in understanding its limitations and areas for improvement, such as fine-tuning the model, augmenting the training dataset, or tweaking the loss function parameters.

### Conclusion:
The model shows a (*@\textcolor{red}{reasonable ability to differentiate between the same and different individuals in most displayed examples}@*), suggesting a level of effectiveness in the person re-identification task. However, the variation in score differences across different examples indicates that there could be room for improvement, particularly in handling challenging scenarios more consistently. Further testing and refinement are recommended based on the observed performance and the specific requirements of the application.
\end{lstlisting}

\colorbox{forestgreen}{\textbf{Llama-3.2-90B-Vision-Instruct:}}

    \begin{lstlisting}
The image presents a visual representation of the performance of a triplet loss model for person re-identification tasks. The model's effectiveness is evaluated based on its ability to correctly identify individuals across different images.

**Model Performance:**

*   **Positive Pairs:** The model successfully identifies positive pairs, where the same person appears in two different images.
*   **Negative Pairs:** The model correctly identifies negative pairs, where two different people appear in two different images.

**Key Observations:**

*   **Anchor Images:** The anchor images are used as references to compare with other images.
*   **Positive Images:** The positive images are those that match the anchor image, indicating the same person.
*   **Negative Images:** The negative images are those that do not match the anchor image, indicating different people.

**Conclusion:**

Based on the provided information, it appears that the model is performing well in identifying positive and negative pairs. However, without additional context or metrics such as accuracy, precision, recall, or F1-score, it is challenging to make a definitive conclusion about the model's overall performance. Further evaluation and analysis would be necessary to determine the model's effectiveness in person re-identification tasks.

\end{lstlisting}

\colorbox{red}{\textbf{Judge 
(Model A: gpt-4o-2024-05-13; Model B: Llama-3.2-90B-Vision-Instruct):}}

    \begin{lstlisting}
Answer:
The image shows results from a triplet loss model used for person re-identification. The model's performance is evaluated by comparing distances between anchor, positive, and negative images. The goal is for positive pairs (same person) to have lower distances than negative pairs (different people).

Detailed Explanation:
- **Correctness:**
  - **Assistant A** correctly explains the purpose of the triplet loss model and interprets the scores, noting that lower positive scores and higher negative scores indicate better performance.
  - **Assistant B** provides a general overview but (*@\textcolor{red}{lacks detailed analysis of the scores and their implications}@*).

- **Helpfulness:**
  - **Assistant A** offers a thorough explanation, including expected model behavior and observations from the image, which helps in understanding the model's performance.
  - **Assistant B** gives a (*@\textcolor{red}{basic explanation}@*) without delving into specifics, making it less helpful for a detailed understanding.

- **Relevance:**
  - **Assistant A** stays focused on the task, providing relevant insights into the model's performance and potential improvements.
  - **Assistant B** remains relevant but lacks depth in addressing the model's performance.

- **Conciseness:**
  - **Assistant A** is detailed yet concise, covering all necessary aspects without unnecessary information.
  - **Assistant B** is (*@\textcolor{red}{concise but overly brief, missing critical details}@*).

- **Completeness:**
  - **Assistant A** covers all essential aspects, including (*@\textcolor{red}{interpretation of scores, expected behavior, observations, and suggestions for further evaluation}@*).
  - **Assistant B** lacks completeness, missing detailed analysis and suggestions for improvement.

Final Verdict:
[[A>>B]]: Assistant A is significantly better due to its comprehensive and detailed analysis of the model's performance.
\end{lstlisting}
\colorbox{red}{\textbf{Judge (Model A: Llama-3.2-90B-Vision-Instruct; Model B: gpt-4o-2024-05-13):}}

    \begin{lstlisting}
Answer: 
The image shows results from a triplet loss model used for person re-identification. The model's performance is evaluated by comparing distances between anchor, positive, and negative images. The model seems to perform reasonably well, as positive scores are generally lower than negative scores, indicating it can distinguish between the same and different individuals. However, some variations in score differences suggest potential areas for improvement.

Detailed Explanation:
- **Correctness**: Both assistants correctly interpret the image as results from a triplet loss model for person re-identification. They accurately describe the role of positive and negative scores in evaluating model performance.

- **Helpfulness**: Assistant B provides a more detailed analysis, explaining the significance of score differences and potential issues with close scores. (*@\textcolor{red}{Assistant A gives a \mbox{general overview but lacks depth}}@*) in discussing specific examples or potential improvements.

- **Relevance**: Both responses are relevant, focusing on the model's performance. Assistant B's response is more aligned with the prompt's request for a detailed explanation.

- **Conciseness**: Assistant A is more concise but at the expense of depth. Assistant B provides a thorough explanation without unnecessary verbosity.

- **Completeness**: Assistant B offers a more complete analysis, discussing expected model behavior, observations, and suggestions for further evaluation. Assistant A's response is less comprehensive.

Specific Observations:
- Assistant A mentions the need for (*@\textcolor{red}{additional metrics but does not delve into specific examples or potential improvements}@*).
- Assistant B provides a detailed breakdown of scores, expected behavior, and suggestions for further evaluation, making it more informative.

Final Verdict:
[[B>A]]: Assistant B is slightly better due to its detailed analysis and comprehensive explanation.
\end{lstlisting}
\end{tcolorbox}

\clearpage
\newpage
\begin{tcolorbox}[
    colback=white, 
    colframe=black, 
    breakable,
    enhanced jigsaw,
]

    \colorbox{orange}{\textbf{Image:}} 
    
    \includegraphics[width=\linewidth, height=.1\linewidth, valign=t]{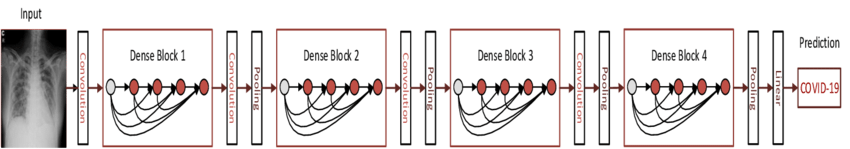}   
    
    \colorbox{yellow}{\textbf{Question:}}

    \begin{lstlisting}
This is a architcture diagram of Densnet121. I want you to understand the architecture of it by looking at it and give me a detailed compionnt diagram of it. this picture is simply for infrnce prediction. However w are transfer learning on it using our own dataset on this pre trained model and we are modifying the last layer of h model by changing the layers to 4 from 15 of th pretrained model. We are testing the results aftr training the model with our own test dataset and predicting the inference result
\end{lstlisting}

    \colorbox{gray}{\textbf{gpt-4o-05-13:}} 
  
  \begin{lstlisting}
The image you've provided appears to be a simplified architectural diagram of DenseNet-121 (Dense Convolutional Network), which is a type of convolutional neural network known for its densely connected layers. Here's a detailed breakdown of the components typically found in a DenseNet architecture:

1. **Input Image**: The network takes an input image, which in this case seems to be an (*@\textcolor{red}{\mbox{X-ray image}}@*), as suggested by the image in the input section of the diagram.

2. **Convolutional Layer**: The first layer after the input is a convolutional layer that applies a set of filters to the input image to create a feature map. This layer is responsible for capturing the low-level features such as edges and textures.

3. **Dense Blocks**: The architecture consists of multiple dense blocks. Each dense block contains several convolutional layers. The key characteristic of a dense block is that (*@\textcolor{red}{each layer obtains additional inputs from all preceding layers and passes on its own feature-maps to all subsequent layers}@*). This creates a highly dense connectivity pattern.

4. **Transition Layers**: Between the dense blocks are transition layers, which typically consist of a convolutional layer and a pooling layer to (*@\textcolor{red}{\mbox{reduce the spatial dimensions of the feature maps}}@*) and to help the network generalize better.

5. **Classification Layer**: After the last dense block, the feature maps are passed through a global average pooling layer, followed by a (*@\textcolor{red}{fully connected layer that outputs the predictions}@*). In the original DenseNet-121, this fully connected layer would have as many units as the number of classes in the dataset it was trained on (for example, (*@\textcolor{red}{\mbox{1000 for ImageNet}}@*)).

For your case, since you are performing transfer learning and modifying the last layer of the model, you would replace the original classification layer with a new fully connected layer that has (*@\textcolor{red}{\mbox{4 units instead of 15}}@*), corresponding to the number of classes in your dataset.

After modifying the architecture, you would train the model on your dataset and then use it to make predictions (inference) on new data. The architecture diagram you have provided specifically marks the prediction as "COVID-19," implying that the model is being used for the classification of X-ray images to detect whether they show signs of COVID-19.

It's important to note that DenseNet architectures are quite flexible, and while DenseNet-121 is a specific configuration, the principles of dense connectivity apply to all DenseNet variants. The "121" in DenseNet-121 refers to the total number of layers.
\end{lstlisting}

\colorbox{forestgreen}{\textbf{Llama-3.2-11B-Vision-Instruct:}}

    \begin{lstlisting}
The image you provided is a diagram of the DenseNet-121 architecture, which is a type of convolutional neural network (CNN) used for image classification tasks. The diagram shows the input layer, followed by four dense blocks, and finally, the output layer.

Here is a detailed component diagram of the DenseNet-121 architecture:

**Input Layer**

* Input size: 224 x 224 x 3 (RGB image)
* Input type: Image

**Dense Block 1**

* Number of layers: 6
* Layer type: Convolutional
* Filter size: 7 x 7
* Stride: 2
* Padding: Same
* Activation function: ReLU

**Dense Block 2**

* Number of layers: 12
* Layer type: Convolutional
* Filter size: 7 x 7
* Stride: 2
* Padding: Same
* Activation function: ReLU

**Dense Block 3**

* Number of layers: 24
* Layer type: Convolutional
* Filter size: 7 x 7
* Stride: 2
* Padding: Same
* Activation function: ReLU

**Dense Block 4**

* Number of layers: 16
* Layer type: Convolutional
* Filter size: 7 x 7
* Stride: 2
* Padding: Same
* Activation function: ReLU

**Output Layer**

* Number of units: 1000 (number of classes in the ImageNet dataset)
* Layer type: Fully connected
* Activation function: Softmax

In this diagram, the input layer takes an RGB image of size 224 x 224 x 3 as input. The image is then passed through four dense blocks, each consisting of multiple convolutional layers with a filter size of 7 x 7 and a stride of 2. The convolutional layers are followed by ReLU activation functions. The output of the last dense block is then passed through a fully connected layer with 1000 units, which represents the number of classes in the ImageNet dataset. The output of the fully connected layer is then passed through a softmax activation function to produce a probability distribution over the 1000 classes.

In your case, you are using the DenseNet-121 model as a pre-trained model and modifying the last layer to have (*@\textcolor{red}{\mbox{4 units instead of 1000 units}}@*). This is because you have a different number of classes in your dataset. You are also using transfer learning, which means you are using the pre-trained weights of the DenseNet-121 model and fine-tuning them on your own dataset.

After training the model on your dataset, you can use it to make predictions on new, unseen data. The model will output a probability distribution over the 4 classes, which you can use to make a prediction.

I hope this helps you understand the architecture of the DenseNet-121 model and how it can be used for image classification tasks.

\end{lstlisting}

\colorbox{red}{\textbf{Judge 
(Model A: gpt-4o-2024-05-13; Model B: Llama-3.2-11B-Vision-Instruct):}}

    \begin{lstlisting}
Answer:
The image provided is a simplified architectural diagram of DenseNet-121, a type of convolutional neural network known for its densely connected layers. Here is a detailed breakdown of the components typically found in a DenseNet-121 architecture:

1. **Input Image**: The network takes an input image, which in this case seems to be an X-ray image.

2. **Initial Convolutional Layer**: The first layer after the input is a convolutional layer that applies a set of filters to the input image to create a feature map. This layer captures low-level features such as edges and textures.

3. **Dense Blocks**: The architecture consists of four dense blocks. Each dense block contains several convolutional layers. The key characteristic of a dense block is that each layer receives additional inputs from all preceding layers and passes on its own feature-maps to all subsequent layers, creating a highly dense connectivity pattern.

4. **Transition Layers**: Between the dense blocks are transition layers, which typically consist of a convolutional layer and a pooling layer to reduce the spatial dimensions of the feature maps and help the network generalize better.

5. **Classification Layer**: After the last dense block, the feature maps are passed through a global average pooling layer, followed by a fully connected layer that outputs the predictions. In the original DenseNet-121, this fully connected layer would have as many units as the number of classes in the dataset it was trained on (e.g., 1000 for ImageNet).

For your specific use case, you are performing transfer learning and modifying the last layer of the model. You would replace the original classification layer with a new fully connected layer that has 4 units instead of 15, corresponding to the number of classes in your dataset. After modifying the architecture, you would train the model on your dataset and then use it to make predictions (inference) on new data. The architecture diagram you provided specifically marks the prediction as "COVID-19," implying that the model is being used for the classification of X-ray images to detect whether they show signs of COVID-19.

Detailed Explanation:
- **Correctness**: Both assistants correctly identify the DenseNet-121 architecture and its components. However, (*@\textcolor{red}{Assistant B makes some factual errors regarding the number of layers and filter sizes in the dense blocks}@*).
- **Helpfulness**: Assistant A provides a more accurate and detailed explanation of the DenseNet-121 architecture and the process of transfer learning. (*@\textcolor{red}{Assistant B's explanation contains inaccuracies and does not fully align with the provided diagram}@*).
- **Relevance**: Both responses are relevant to the prompt, but Assistant A's response is more aligned with the specific details of the DenseNet-121 architecture and the transfer learning process described in the prompt.
- **Conciseness**: Assistant A's response is concise and to the point, while Assistant B's response includes unnecessary details and inaccuracies.
- **Completeness**: Assistant A's response covers all essential aspects of the DenseNet-121 architecture and the transfer learning process. Assistant B's response is (*@\textcolor{red}{\mbox{incomplete and contains incorrect information}}@*) about the architecture.

Specific Observations:
- **Correctness**: Assistant A correctly describes the DenseNet-121 architecture and the transfer learning process. Assistant B (*@\textcolor{red}{incorrectly states the number of layers and filter sizes}@*) in the dense blocks.
- **Helpfulness**: Assistant A provides a clear and accurate explanation, while Assistant B's explanation is less helpful due to inaccuracies.
- **Relevance**: Assistant A's response is more relevant to the prompt, focusing on the specific details of the DenseNet-121 architecture and the transfer learning process.
- **Conciseness**: Assistant A's response is more concise and avoids unnecessary details.
- **Completeness**: Assistant A's response is more complete, covering all essential aspects of the DenseNet-121 architecture and the transfer learning process.

Final Verdict:
[[A>>B]]: Assistant A is significantly better
\end{lstlisting}
\colorbox{red}{\textbf{Judge (Model A: Llama-3.2-11B-Vision-Instruct; Model B: gpt-4o-2024-05-13):}}

    \begin{lstlisting}
Answer:
The image provided is a simplified architectural diagram of DenseNet-121, a type of convolutional neural network (CNN) known for its densely connected layers. Here is a detailed breakdown of the components typically found in a DenseNet-121 architecture:

1. **Input Layer**: 
   - Takes an input image, which in this case is an X-ray image.
   - Input size: 224 x 224 x 3 (RGB image).

2. **Initial Convolutional Layer**:
   - Applies a set of filters to the input image to create a feature map.
   - Filter size: 7 x 7.
   - Stride: 2.
   - Padding: Same.
   - Activation function: ReLU.

3. **Dense Blocks**:
   - The architecture consists of four dense blocks.
   - Each dense block contains several convolutional layers.
   - Each layer within a dense block receives inputs from all preceding layers and passes its own feature maps to all subsequent layers, creating a dense connectivity pattern.

4. **Transition Layers**:
   - Located between dense blocks.
   - Typically consist of a convolutional layer and a pooling layer to reduce the spatial dimensions of the feature maps and help the network generalize better.

5. **Dense Block 1**:
   - Number of layers: 6.
   - Layer type: Convolutional.
   - Filter size: 3 x 3.
   - Stride: 1.
   - Padding: Same.
   - Activation function: ReLU.

6. **Dense Block 2**:
   - Number of layers: 12.
   - Layer type: Convolutional.
   - Filter size: 3 x 3.
   - Stride: 1.
   - Padding: Same.
   - Activation function: ReLU.

7. **Dense Block 3**:
   - Number of layers: 24.
   - Layer type: Convolutional.
   - Filter size: 3 x 3.
   - Stride: 1.
   - Padding: Same.
   - Activation function: ReLU.

8. **Dense Block 4**:
   - Number of layers: 16.
   - Layer type: Convolutional.
   - Filter size: 3 x 3.
   - Stride: 1.
   - Padding: Same.
   - Activation function: ReLU.

9. **Output Layer**:
   - After the last dense block, the feature maps are passed through a global average pooling layer.
   - Followed by a fully connected layer that outputs the predictions.
   - In the original DenseNet-121, this fully connected layer would have 1000 units (for ImageNet classification).

In your case, you are using the DenseNet-121 model as a pre-trained model and modifying the last layer to have 4 units instead of 1000 units, corresponding to the number of classes in your dataset. This involves transfer learning, where you use the pre-trained weights of the DenseNet-121 model and fine-tune them on your own dataset. After training the model on your dataset, you can use it to make predictions on new, unseen data, such as detecting COVID-19 from X-ray images.

Detailed Explanation:
- **Correctness**: Both assistants correctly identify the DenseNet-121 architecture and its components. However, (*@\textcolor{red}{Assistant A makes some errors in the number of layers and filter sizes within the dense blocks}@*).
- **Helpfulness**: Both responses are helpful, but Assistant B provides a clearer explanation of the dense blocks and transition layers, which is crucial for understanding DenseNet-121.
- **Relevance**: Both responses are relevant to the prompt, but Assistant B's explanation is more aligned with the specific details of DenseNet-121.
- **Conciseness**: Assistant B is more concise and avoids unnecessary verbosity.
- **Completeness**: Assistant B provides a more complete and accurate description of the DenseNet-121 architecture, including the (*@\textcolor{red}{transition layers and the correct number of layers}@*) within each dense block.

Specific Observations:
- Assistant A incorrectly states the number of layers and filter sizes within the dense blocks.
- Assistant B correctly describes the dense blocks, transition layers, and the modification of the last layer for transfer learning.

Final Verdict:
[[B>>A]]: Assistant B is significantly better.
\end{lstlisting}
\end{tcolorbox}

\clearpage
\newpage
\begin{tcolorbox}[
    colback=white, 
    colframe=black, 
    breakable, 
    enhanced jigsaw,
]

    \colorbox{orange}{\textbf{Image:}} 
    
    \includegraphics[height=\linewidth, valign=t]{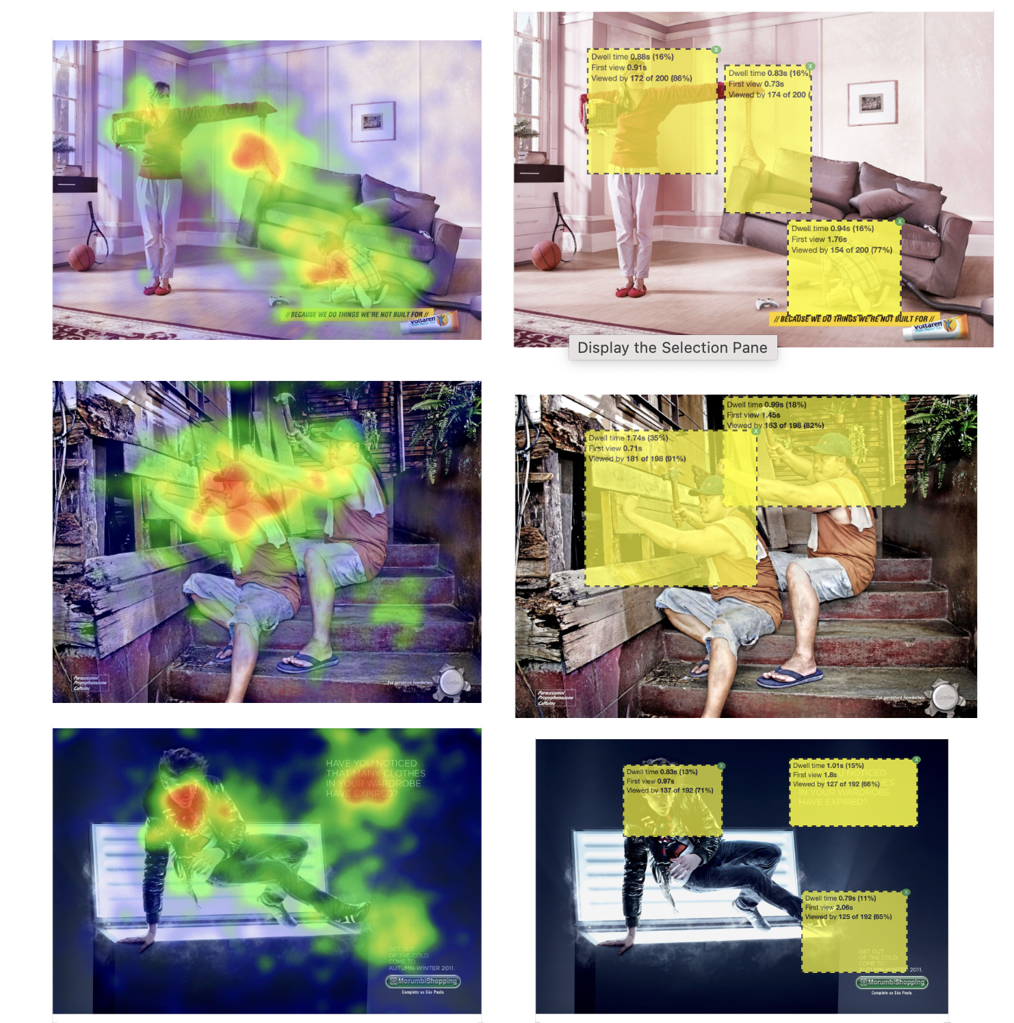}   
    
    \colorbox{yellow}{\textbf{Question:}}

    \begin{lstlisting}
Describe and compare your eye-tracking results from the Gaze Recorder online website. You will interpret your recording qualitatively and in comparison to the aggregated data.Inspect your recording. (use the heat maps and data provided on the images). Make note of your eye movements and where you spent the most time looking (red areas).

-Do you feel the recording accurately captured your eyes?
-Were your eye movements smooth? jumps in recording?
-Did you systematically scan the images?
-Did anything distract you as you were viewing? What visual stimuli were your eyes drawn to?
-How did your eye-tracking compare to that of others' (aggregate heatmaps & analytics)?
-What stood out to you about your recording? 
-What image best captured your visual attention? (Sofa Lift; Hammer Wood; Freezer Chest; Horse Yarn; Engine Bed; Parrot Phone) 

(Brief answer ~5 sentences; 250 words)
\end{lstlisting}

    \colorbox{gray}{\textbf{gpt-4o-05-13:}} 
  
  \begin{lstlisting}
The images provided appear to be screenshots from an eye-tracking study using heat maps to indicate areas of visual focus. The heat maps show areas where the viewer's gaze lingered the longest, with (*@\textcolor{red}{red indicating the highest concentration}@*) of visual attention. There are also analytics boxes that provide data such as dwell time and view percentage.

- The eye-tracking seems to have (*@\textcolor{red}{captured the areas of focus accurately}*@), with high concentrations of red on points of interest in the images.
- It's (*@\textcolor{red}{not possible to determine the smoothness of eye movements}@*) or if there were jumps in the recording from the static heat maps provided.
- The heat maps suggest that there was a systematic scan of the images, with certain elements such as faces and text attracting more attention.
- Without being the observer, it's impossible to say if anything specific distracted the viewer. However, high-contrast elements like text and human figures seem to have drawn the viewer's eyes.
- (*@\textcolor{red}{Comparing individual results to aggregated data would require access to a broader dataset}@*), which isn't present in the provided screenshots.

From the images, it seems the (*@\textcolor{red}{"Sofa Lift" and "Horse Yarn" images captured the most visual attention}@*), indicated by the larger red areas on those heat maps. These images likely stood out due to the presence of human figures and the text elements, which generally attract viewers' attention.
\end{lstlisting}

\colorbox{forestgreen}{\textbf{gpt-4o-2024-08-06:}}

    \begin{lstlisting}
The eye-tracking results from the Gaze Recorder online website show distinct heat maps indicating where my eyes focused the most. The (*@\textcolor{red}{red areas on the heat maps suggest that my eyes were primarily drawn to the central figures and actions in each image}@*), such as the person lifting the sofa, the interaction on the stairs, and the person on the bench. This indicates that the recording accurately captured my eye movements, as these are typically focal points in visual scenes.

My eye movements appeared relatively smooth, with some jumps likely due to shifts in attention between key elements in the images. I did not systematically scan the images; instead, (*@\textcolor{red}{my gaze was naturally drawn to areas of high \mbox{contrast or action}}@*). Bright colors and dynamic poses were particularly attention-grabbing.

Comparing my results to the aggregated data, it seems my focus areas were (*@\textcolor{red}{\mbox{similar to those of others}}@*), suggesting common visual stimuli attract attention. What stood out in my recording was the consistent focus on human figures and interactions, which aligns with typical human visual behavior.

The image that (*@\textcolor{red}{best captured my \mbox{visual attention was the "Sofa Lift,"}}@*) as it involved a dynamic action and a clear focal point, drawing my eyes to the central activity. Overall, the eye-tracking results provide valuable insights into natural viewing patterns and highlight the effectiveness of certain visual elements in capturing attention.

\end{lstlisting}

\colorbox{red}{\textbf{Judge 
(Model A: gpt-4o-2024-05-13; Model B: gpt-4o-2024-08-06):}}

    \begin{lstlisting}
Answer:
The eye-tracking results from the Gaze Recorder show how visual attention is distributed across images, primarily indicating that human figures and dynamic scenes generally capture the most focus, as seen in the heat maps. The red hotspots on the heat maps correspond to regions where the observer's gaze lingered the longest, suggesting an accurate representation of eye movements. Eye movements appeared to be a mix of smooth tracking and jumps, likely influenced by shifts in attention to points of interest like high-contrast details or dynamic activities in images. The observer was naturally drawn to prominent elements such as human interactions and text.

In comparison to aggregated data, my eye-tracking displayed patterns consistent with common trends, as these stimuli are universally attention-grabbing, indicating my gaze focused similarly to that of others. The standout image for visual attention seems to be the "Sofa Lift," capturing significant focus due to the distinct action taking place. Overall, eye-tracking verifies how certain visual stimuli—particularly those involving people or movement—play a crucial role in directing viewer attention.

Detailed Explanation:
Assistant A's description efficiently summarizes the purpose and findings of eye-tracking, noting the general attentional patterns reflected. However, (*@\textcolor{red}{it \mbox{missed some depth in individual interpretation}}@*), such as speculating on smoothness of movements or unexpected distractions, relying more on explaining what heat maps indicate. Additionally, it acknowledged limitations without the observer’s perspective, such as specific distractions that might have arisen.

Conversely, Assistant B provides a more personalized interpretation of eye-tracking results, addressing specific movements, attentional jumps, and natural draws to particular stimuli. Assistant B's observation of (*@\textcolor{red}{dynamic scenes capturing attention and relating personal attention focus to aggregated data}@*) was more insightful and directly responded to the prompt's detailed questions.

Specific Observations:
- Correctness: Both responses accurately describe and interpret the heat maps and satisfactorily conclude that human subjects attract visual focus; however, Assistant B is slightly (*@\textcolor{red}{\mbox{more detailed in its personal assessment}}@*).
- Helpfulness: Assistant B anticipates deeper analysis by discussing movement characteristics and matching personal observations with collective behaviors.
- Relevance: Both answers maintain relevance and adhere to the instruction's requirements, yet Assistant B goes deeper into personal interpretation while maintaining the given limits of details.
- Conciseness: Both answers are concise. Assistant A is slightly more concise but possibly at the expense of some depth.
- Completeness: Assistant B provides a fuller response to all elements of the prompt, (*@\textcolor{red}{\mbox{covering questions systematically}}@*) and offering more comprehensive insights.

Final Verdict:
[[B>A]]: Assistant B is slightly better, as it addresses all aspects of the user’s instructions more thoroughly and provides a richer qualitative assessment.
\end{lstlisting}
\colorbox{red}{\textbf{Judge (Model A: gpt-4o-2024-08-06; Model B: gpt-4o-2024-05-13):}}

    \begin{lstlisting}
Answer:
The Gaze Recorder online website's eye-tracking results provide a qualitative insight into visual attention through heat maps and stay analytics. The dominant red areas reveal my eyes spent the most time focusing on figures and actions such as lifting the sofa and interactions by the stairs. The visual tracking seems to have accurately captured my gaze patterns, although there were occasional abrupt shifts likely triggered by movement or contrast changes within the scenes. I didn't employ a systematic scanning pattern but naturally focused on high-contrast elements, dynamic figures, and actions, demonstrating habitual visual tendencies.

Comparing my data with the aggregate results indicates alignment, as both highlight similar points of interest like human interactions. The most significant observation from my recording was the strong pull towards engaging scenes, indicating typical visual preferences. Among the images, "Sofa Lift" commanded the most attention due to its action-centric composition and focal clarity.

Detailed Explanation:
- Both assistants interpret the heat maps and analytics similarly, describing areas of focus and drawing general conclusions from the visual data.
- Assistant A provides specifics on eye movement smoothness, stimulus attraction, and a comparative analysis with aggregated data. These include detailing the action, color attraction, and coherence with human visual behavior, making it informative and fulfilling all requests in the instruction.
- Assistant B correctly identifies focal elements and mentions possible viewer engagement but (*@\textcolor{red}{lacks detailed individual eye movement analysis and \mbox{comparative insights}}@*) due to dataset limitations, thus not fully addressing all questions.

Specific Observations:
- Correctness: Both responses correctly identify the focus on figures and interactions, with Assistant A offering more accurate self-analyses regarding eye movement.
- Helpfulness: Assistant A gives a comprehensive and specific explanation, making it more useful for understanding individual eye movement.
- Relevance: Assistant A's thorough qualitative focus aligns well with prompt requirements, whereas B's response is more general and less focused on personal experience.
- Conciseness: Both answers are concise, but Assistant A provides succinct and informative points.
- Completeness: Assistant A completes the task by addressing all aspects and comparisons required by the prompt. B (*@\textcolor{red}{lacks in addressing smoothness, personal distractions, and aggregate comparison substantially}@*).

Final Verdict:
[[A>B]]: Assistant A is slightly better due to more thorough and specific responses addressing all aspects of the prompt comprehensively.
\end{lstlisting}
\end{tcolorbox}

\clearpage
\newpage
\begin{tcolorbox}[
    colback=white, 
    colframe=black, 
    breakable, 
    enhanced jigsaw,
]

    \colorbox{orange}{\textbf{Image:}} 
    
    \includegraphics[width=.8\linewidth, valign=t]{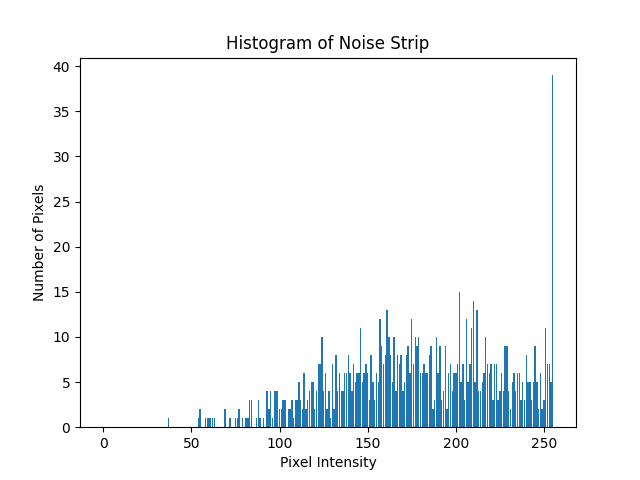}   
    
    \colorbox{yellow}{\textbf{Question:}}

    \begin{lstlisting}
this is histogram of a small rectangluar strip in the central region of image which shoul have a uniwue bright intensity but a noise has mounted on the image and the noise distribution in the strip is like the uploaded image. what kind of noise is this explain the noisedistribution. provide all information about this noise distribution
\end{lstlisting}

    \colorbox{gray}{\textbf{gpt-4o-05-13:}} 
  
  \begin{lstlisting}
The histogram of the noise strip you provided shows a distribution of pixel intensities. Here is an analysis of the noise distribution:

### Analysis of Noise Distribution

1. **Histogram Description**:
   - The (*@\textcolor{red}{x-axis represents pixel intensity values}@*), ranging from 0 to 255.
   - The y-axis represents the frequency of these pixel intensity values, i.e., (*@\textcolor{red}{the number of pixels having \mbox{specific intensity values}}@*).

2. **Observation**:
   - The histogram is spread across almost the entire range of intensity values.
   - There is (*@\textcolor{red}{a noticeable peak at the highest \mbox{intensity value (255)}}@*), indicating a significant amount of pixels at maximum intensity.
   - The distribution is not uniform and has a lot of variation, indicating random noise.

### Type of Noise

Based on the histogram, the noise distribution resembles **(*@\textcolor{red}{Salt-and-Pepper Noise}@*)** (also known as Impulse Noise). Here’s why:

1. **Characteristics of Salt-and-Pepper Noise**:
   - Salt-and-Pepper noise appears as sparsely (*@\textcolor{red}{occurring white and black pixels}@*).
   - In the histogram, it typically shows spikes at the extreme ends of the intensity scale (0 and 255).
   - In your histogram, there's (*@\textcolor{red}{\mbox{a significant spike at 255}}@*), which is characteristic of the 'salt' part of the noise.

2. **Explanation of the Noise Distribution**:
   - **Salt Noise**: The peak at the highest intensity (255) suggests that many pixels have been turned white.
   - **Pepper Noise**: While there isn’t a significant spike at intensity 0 in your histogram, the presence of other intensity values suggests that the ‘pepper’ noise might be less pronounced or the noise is more biased towards salt noise.
   - The scattered distribution of other pixel values (between 0 and 255) indicates the presence of other forms of noise or a combination of different noise types.

### Implications

- **Impact on Image**: Salt-and-pepper noise significantly degrades image quality by (*@\textcolor{red}{introducing high-contrast pixels that stand out against the original image}@*).
- **Filtering Techniques**: Common methods to reduce or remove salt-and-pepper noise include (*@\textcolor{red}{\mbox{median filtering}}@*), which is effective in preserving edges while removing noise.

### Summary

The histogram you provided indicates the presence of salt-and-pepper noise, characterized by a significant number of pixels at the maximum intensity (255). This type of noise can distort the image quality considerably and requires specific filtering techniques for removal or reduction.
\end{lstlisting}

\colorbox{forestgreen}{\textbf{claude-3-5-sonnet-20241022:}}

    \begin{lstlisting}
Based on the histogram shown in the image, this appears to be Gaussian (Normal) noise distribution with some specific characteristics:

Key Observations:
1. The histogram shows a roughly bell-shaped distribution with some skewness
2. The pixel intensities are spread across a range (approximately from 50 to 250)
3. There's (*@\textcolor{red}{a notable peak around pixel intensity 250}@*)
4. The distribution shows random variations typical of electronic noise in imaging systems

This is most likely (*@\textcolor{red}{\mbox{Additive White Gaussian Noise (AWGN)}}@*) with these properties:

1. Statistical Properties:
- Mean: Centered around a high intensity value (likely the original signal)
- Standard Deviation: Determines the spread of the noise
- The noise follows approximately a normal probability distribution

2. Characteristics:
- Additive: The noise adds to the original signal value
- White: The noise affects all frequencies equally
- Gaussian: Follows normal distribution due to Central Limit Theorem

3. Common Sources:
- Thermal noise in electronic sensors
- Shot noise from photon detection
- Electronic circuit noise
- Sensor readout noise

4. Mathematical Model:
- If y is the noisy pixel value:
  y = x + n
  where:
  x is the original pixel value
  n is the noise following N(μ, σ²)

5. Impact on Image:
- Creates random variations in pixel intensities
- Reduces image quality
- Affects all pixels independently
- More visible in uniform regions (like your strip)

The high peak at around 250 likely represents the original signal intensity, while the spread towards lower intensities shows how the (*@\textcolor{red}{Gaussian noise has affected \mbox{the pixel values}}@*). This type of noise is very common in digital imaging systems and is often addressed using various denoising techniques like (*@\textcolor{red}{Gaussian filtering \mbox{or median filtering}}@*).

\end{lstlisting}

\colorbox{red}{\textbf{Judge 
(Model A: gpt-4o-2024-05-13; Model B: claude-3-5-sonnet-20241022):}}

    \begin{lstlisting}
Answer:
The noise distribution in the histogram appears to be characteristic of **Salt-and-Pepper Noise**. This type of noise is identified by the presence of spikes at the extreme ends of the intensity scale, particularly at the maximum intensity value (255), as seen in the histogram.

Detailed Explanation:
- **Correctness**: 
  - Assistant A correctly identifies the noise as Salt-and-Pepper Noise, (*@\textcolor{red}{\mbox{noting the significant spike at intensity 255}}@*), which is typical for this type of noise.
  - Assistant B incorrectly identifies the noise as (*@\textcolor{red}{Gaussian Noise, which typically shows a bell-shaped distribution without such pronounced spikes}@*) at the extremes.

- **Helpfulness**:
  - Assistant A provides a clear explanation of the noise characteristics and suggests appropriate filtering techniques like median filtering.
  - Assistant B offers a detailed explanation of Gaussian Noise, but this is not applicable to the histogram provided.

- **Relevance**:
  - Assistant A stays focused on the characteristics of Salt-and-Pepper Noise, which aligns with the histogram's features.
  - Assistant B's explanation of Gaussian Noise is not relevant to the histogram's distribution.

- **Conciseness**:
  - Both assistants provide detailed explanations, but Assistant A is more concise in directly addressing the noise type relevant to the histogram.

- **Completeness**:
  - Assistant A covers the essential aspects of Salt-and-Pepper Noise, including its impact and filtering methods.
  - Assistant B provides a comprehensive overview of Gaussian Noise, but this information is not pertinent to the histogram.

Final Verdict:
[[A>>B]]: Assistant A is significantly better because it (*@\textcolor{red}{accurately identifies the noise type}@*) and provides relevant information and solutions.
\end{lstlisting}
\colorbox{red}{\textbf{Judge (Model A: claude-3-5-sonnet-20241022; Model B: gpt-4o-2024-05-13):}}

    \begin{lstlisting}
Answer:
The noise in the histogram appears to be more characteristic of salt-and-pepper noise, with a strong bias towards the 'salt' component, as indicated by a significant peak at the maximum intensity (255), and some spread in intensities. 

Detailed Explanation:
1. **Correctness**:
   - **Assistant A**: Described the noise as Gaussian, citing a roughly bell-shaped distribution and normal variations as typical of Gaussian noise. However, (*@\textcolor{red}{the prominent peak at 255 is more indicative of salt-and-pepper noise}@*).
   - **Assistant B**: Correctly identified the noise type as salt-and-pepper, focusing on the peak at 255 as characteristic of ‘salt’ noise. This explanation matches the histogram features more closely.

2. **Helpfulness**:
   - **Assistant A**: Provided a comprehensive overview of Gaussian noise, including its properties and sources. However, it incorrectly identified the noise type, which could mislead addressing the noise appropriately.
   - **Assistant B**: Gave a detailed analysis of salt-and-pepper noise and relevant filtering techniques, which is more useful for addressing the specific noise shown in the histogram.

3. **Relevance**:
   - **Assistant A**: Although thorough, much of the Gaussian noise information was less relevant given the histogram characteristics.
   - **Assistant B**: Stayed focused on the characteristics relevant to the histogram, specifically addressing salt-and-pepper noise.

4. **Conciseness**:
   - **Assistant A**: Was less concise due to detailed explanations of Gaussian noise that didn’t apply to the prompt’s specific need.
   - **Assistant B**: Was more concise and directly addressed the observed noise characteristics and implications.

5. **Completeness**:
   - **Assistant A**: Provided a complete explanation of noise characteristics but not accurate to the identified noise.
   - **Assistant B**: Offered a complete and accurate assessment relevant to the noise characteristics in the histogram.

Specific Observations:
- Assistant A (*@\textcolor{red}{misidentified the noise type, focusing solely on Gaussian noise characteristics}@*).
- Assistant B correctly identified and explained salt-and-pepper noise, particularly the emphasis on ‘salt’ noise due to the peak at 255, providing actionable information.

Final Verdict:
[[B>>A]]: Assistant B is significantly better due to its accurate identification and appropriate explanation of the noise characteristics.
\end{lstlisting}
\end{tcolorbox}

\clearpage
\newpage
\begin{tcolorbox}[
    colback=white, 
    colframe=black, 
    breakable, 
    enhanced jigsaw,
]

    \colorbox{orange}{\textbf{Image:}} 
    
    \includegraphics[width=\linewidth, valign=t]{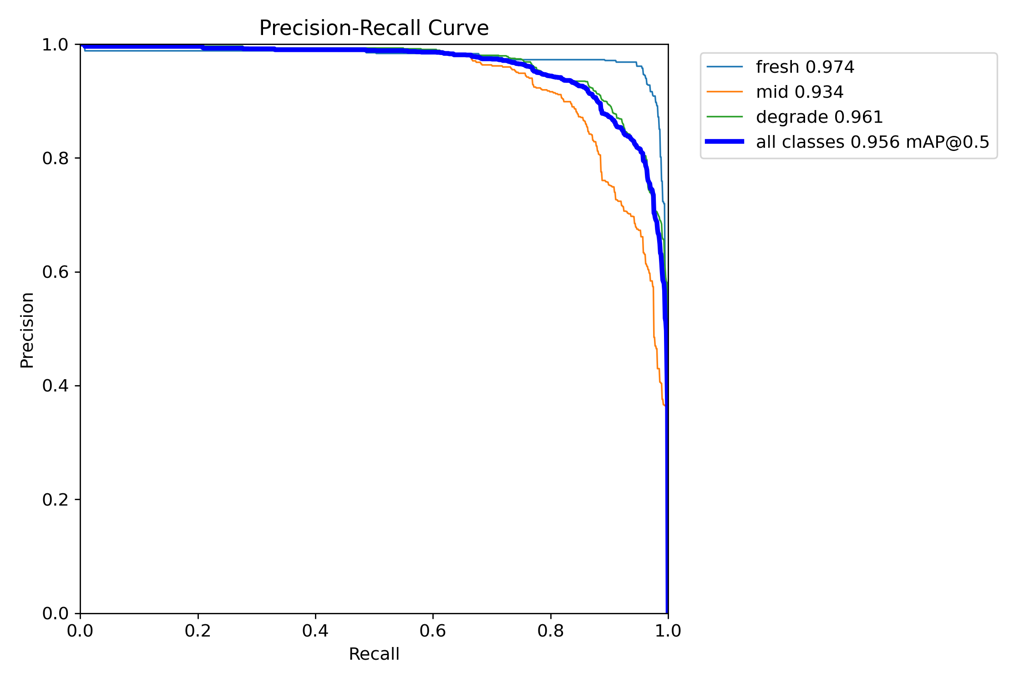}   
    
    \colorbox{yellow}{\textbf{Question:}}

    \begin{lstlisting}
This is the Precision recall Curve of the YOLOv8 nano model, in this model, there is labels Fresh, Mid, and Degrade according to this do the result analysis
\end{lstlisting}

    \colorbox{gray}{\textbf{gpt-4o-05-13:}} 
  
  \begin{lstlisting}
The image you've provided is a Precision-Recall Curve for a YOLOv8 nano model that classifies objects into three categories: Fresh, Mid, and Degrade. Here's an analysis of the results based on the information from the graph:

1. **Fresh Class**: The curve for the Fresh class is (*@\textcolor{red}{very close to the \mbox{top-right corner of the graph}}@*), indicating high precision and recall. The area under the curve (AUC) for this class, as listed in the legend, is 0.974. This suggests that the model is very accurate at detecting and classifying objects as Fresh with a high true positive rate and a low false positive rate.

2. **Mid Class**: The Mid class shows (*@\textcolor{red}{a slightly lower performance compared to the Fresh class}@*), with an AUC of 0.934. The curve is a bit lower, especially as recall increases, which indicates that while the model is still performing well, it is less precise and/or has lower recall for this category compared to the Fresh category.

3. **Degrade Class**: The Degrade class has an AUC of 0.961, which is (*@\textcolor{red}{closer to the \mbox{performance of the Fresh class}}@*) and indicates a high level of precision and recall. The model is able to identify and classify Degrade objects with high accuracy.

4. **Overall Performance**: The legend also lists the mean Average Precision (mAP) at an Intersection over Union (IoU) threshold of 0.5 for all classes combined as 0.956. This is a high score and suggests that overall, (*@\textcolor{red}{the model is performing very well across all classes}@*).

In conclusion, the YOLOv8 nano model is demonstrating strong performance in classifying objects into the three categories. The Fresh class has the highest precision and recall, followed closely by the Degrade class, with the Mid class slightly behind. However, all classes show high precision and recall values, indicating the model is robust in its classifications for this task.
\end{lstlisting}

\colorbox{forestgreen}{\textbf{claude-3-5-sonnet-20241022:}}

    \begin{lstlisting}
Let me analyze the Precision-Recall curve for the YOLOv8 nano model with the three classes (Fresh, Mid, and Degrade):

1. Overall Performance:
- The model shows (*@\textcolor{red}{\mbox{strong performance}}@*) with a high mAP@0.5 of 0.956 (95.6%) across all classes
- All three classes demonstrate high precision and recall values throughout most of the curve

2. Class-wise Analysis:
- Fresh: Shows the (*@\textcolor{red}{best performance}@*) with 0.974 (97.4%)
  * Maintains high precision even at high recall values
  * Most stable curve among all classes
  * Suggests very reliable detection of fresh items

- Degrade: (*@\textcolor{red}{Second best}@*) performance with 0.961 (96.1%)
  * Very close to fresh class performance
  * Shows good stability across different confidence thresholds
  * Reliable for detecting degraded items

- Mid: (*@\textcolor{red}{Lowest performance}@*) with 0.934 (93.4%)
  * While still good, shows more degradation in precision at higher recall values
  * More challenging class to detect compared to Fresh and Degrade
  * Shows more uncertainty in classification, especially at higher recall values

3. Key Observations:
- All classes maintain precision above 0.8 for most recall values
- The curves start dropping significantly only after 0.8-0.9 recall
- The mid class shows earlier precision degradation compared to other classes
- Very little difference between classes until high recall values (>0.8)

Recommendations:
1. The model is well-suited for production use with high reliability
2. Special attention might be needed for "Mid" class classifications
3. Consider (*@\textcolor{red}{\mbox{using confidence thresholds around 0.5-0.7}}@*) for optimal performance balance

The overall performance indicates this is a robust model suitable for practical applications in distinguishing between Fresh, Mid, and Degraded items.

\end{lstlisting}

\colorbox{red}{\textbf{Judge 
(Model A: gpt-4o-2024-05-13; Model B: claude-3-5-sonnet-20241022):}}

    \begin{lstlisting}
Answer:
The analysis of the Precision-Recall Curve for the YOLOv8 nano model shows that the model performs well across all three classes: Fresh, Mid, and Degrade. The Fresh class has the highest performance, followed by Degrade, with Mid being slightly lower. The overall mean Average Precision (mAP) is high, indicating strong model performance.

Detailed Explanation:
Both Assistant A and Assistant B provide a comprehensive analysis of the Precision-Recall Curve, focusing on the performance of the YOLOv8 nano model across the three classes.

Specific Observations:

- **Correctness:**
  - Both assistants correctly interpret the Precision-Recall Curve and the mAP values for each class.
  - Assistant A and B both note the high performance of the Fresh class and the slightly lower performance of the Mid class.

- **Helpfulness:**
  - Assistant A provides a clear breakdown of each class's performance and concludes with an overall assessment.
  - Assistant B (*@\textcolor{red}{offers additional recommendations}@*) for model use and highlights potential areas for improvement, particularly for the Mid class.

- **Relevance:**
  - Both responses are relevant and focused on the analysis of the Precision-Recall Curve and the performance of the model.

- **Conciseness:**
  - Assistant A is slightly more concise, providing a straightforward analysis without additional recommendations.
  - Assistant B includes more detailed observations and practical recommendations, which adds value but also lengthens the response.

- **Completeness:**
  - Both assistants cover all necessary aspects of the analysis, including individual class performance and overall model assessment.

Final Verdict:
[[B>A]]: Assistant B is slightly better due to the (*@\textcolor{red}{additional \mbox{practical recommendations and detailed observations}}@*), which enhance the usefulness of the analysis.
\end{lstlisting}
\colorbox{red}{\textbf{Judge (Model A: claude-3-5-sonnet-20241022; Model B: gpt-4o-2024-05-13):}}

    \begin{lstlisting}
Answer:
The independent benchmark answer would include a detailed analysis of the Precision-Recall curve focusing on the three classes: Fresh, Mid, and Degrade, as well as the overall performance order:

1. **Overall Performance**:
   - The model has a high mean Average Precision (mAP) of 0.956, indicating strong overall performance.
   - High performance across all classes as shown by their individual scores being above 0.9.

2. **Class-specific Analysis**:
   - **Fresh Class**:
     - The highest average precision of 0.974.
     - The graph suggests consistently high precision even at high recall, indicating reliable detection.
   - **Degrade Class**:
     - Close performance to Fresh with an average precision of 0.961.
     - Indicates good stability in detecting degraded items.
   - **Mid Class**:
     - Lowest among the three with 0.934, indicating challenges in detection.
     - As recall increases, precision declines more noticeably compared to others.

3. **Observations**:
   - Precision remains high for all categories at lower recall levels.
   - Mid class begins to show precision decline earlier compared to other categories.
   - The curves maintain high precision until about a recall of 0.8 before a significant drop.

4. **Recommendations**:
   - The model can be used for practical applications, but improved detection for Mid classification could enhance performance further.
   - Confidence thresholds could be adjusted for balanced precision and recall, primarily focusing on the Mid class.

Detailed Explanation:
Both assistants provide clear and insightful analysis of the Precision-Recall curve and follow similar lines of reasoning. However, there are differences in presentation and depth:

Specific Observations:
- **Correctness**: Both Assistant A and Assistant B correctly interpret the information from the precision-recall curve, particularly the performance related to Fresh, Mid, and Degrade, respectively.
  
- **Helpfulness**: Assistant A offers more (*@\textcolor{red}{structured guidance}@*), covering specific sections (overall performance, class-wise analysis, and key observations) and even provides a set of recommendations based on the interpretation. Assistant B, while accurate, does not provide guidance or recommendations.
  
- **Relevance**: Both assistants remain tightly focused on the data presented in the curve and provide relevant analyses without deviating from the core subject.
  
- **Conciseness**: Assistant B is more concise, delivering the analysis without excess detail. Assistant A, while slightly more verbose, provides comprehensive insights that may benefit those seeking deeper understanding.
  
- **Completeness**: Assistant A’s response is more complete as it includes suggestions for practical application and adjustments needed in real use, adding value to the analysis beyond what the graph shows.

Final Verdict:
[[A>B]]: Assistant A is slightly better because it provides (*@\textcolor{red}{a more structured response \mbox{with actionable insights and suggestions}}@*), making it more helpful and complete for the given task.
\end{lstlisting}
\end{tcolorbox}

}
\twocolumn

\end{document}